\documentclass{article}

\usepackage[final]{colm2026_conference}

\usepackage{microtype}
\usepackage{hyperref}
\usepackage{lineno}

\usepackage{graphicx}
\usepackage{caption}
\usepackage{placeins}

\usepackage{algorithm}
\usepackage{algpseudocode}

\usepackage{amsmath}
\usepackage{amssymb}

\usepackage{booktabs}
\usepackage{multirow}
\usepackage{makecell}
\usepackage{array}
\usepackage{colortbl}

\usepackage{pifont}
\usepackage{xcolor}
\usepackage{wrapfig}
\newcommand{\cmark}{\textcolor{blue}{\ding{51}}}
\newcommand{\xmark}{\textcolor{red}{\ding{55}}}

\usepackage{titlesec}
\titlespacing*{\section}{0pt}{8pt plus 2pt minus 2pt}{4pt plus 1pt minus 1pt}
\titlespacing*{\subsection}{0pt}{6pt plus 2pt minus 2pt}{3pt plus 1pt minus 1pt}

\AtBeginDocument{
  \setlength{\abovedisplayskip}{8pt plus 2pt minus 4pt}
  \setlength{\belowdisplayskip}{8pt plus 2pt minus 4pt}
  \setlength{\abovedisplayshortskip}{4pt plus 2pt minus 2pt}
  \setlength{\belowdisplayshortskip}{4pt plus 2pt minus 2pt}
}

\definecolor{darkblue}{rgb}{0, 0, 0.5}
\hypersetup{colorlinks=true, citecolor=darkblue, linkcolor=darkblue, urlcolor=darkblue}

\title{Batch-wise Adaptive Pruning: Periodic Neuron Activation-Aware Weight Pruning for Language Reasoning Model}

\author{
  Yongmin Kim\thanks{Equal contribution.},
  Shota Takashiro\footnotemark[1],
  Yusuke Iwasawa,
  Takeshi Kojima,
  Yutaka Matsuo \\
  The University of Tokyo
}

\begin{document}

\ifcolmsubmission
\linenumbers
\fi

\maketitle
\makeatletter\@topnum\z@\makeatother 
\suppressfloats[t] 
\graphicspath{{figure/}{../figure/}}

\begin{abstract}
Large Reasoning Models (LRMs) achieve strong performance on complex tasks through extended chain-of-thought generation, but incur substantial computational costs during inference. In production settings, batched inference is essential for high throughput, yet the existing training-free adaptive pruning methods we evaluate severely degrade in this regime. Because a batch must share a single pruning mask, these methods aggregate activations across samples and then apply threshold-based selection; the threshold, calibrated offline on unaggregated activations, no longer matches the aggregated distribution, so the realized sparsity ratio drifts and accuracy on reasoning tasks collapses under batched inference.
In this work, we propose a training-free adaptive pruning method designed specifically for batched inference in LRMs, built on two components.
First, we replace threshold-based selection with \emph{periodic top-$k$ selection} over the aggregated importance scores, which is unaffected by the shift that aggregation induces in the activation distribution, and which runs selection once per update period rather than at every token, preserving the speedup.
Second, based on the observation that important neurons re-fire periodically during long reasoning generation, we introduce an \emph{activation memory} that accumulates importance across update phases so that recurring neurons are retained.
Experiments on diverse reasoning benchmarks demonstrate that our method outperforms the previous state-of-the-art adaptive pruning method by 39.7 percentage points in average accuracy at batch size 4 with 50\% target sparsity on DeepSeek-R1-Distill-Qwen-7B, and reaches $1.40\times$ speedup over dense inference at 50\% actual sparsity. \footnote{Our code is available at: \url{https://github.com/matsuolab/batch-wise-prune}}
\end{abstract}


\begin{figure}[t]
\centering
\includegraphics[width=\linewidth]{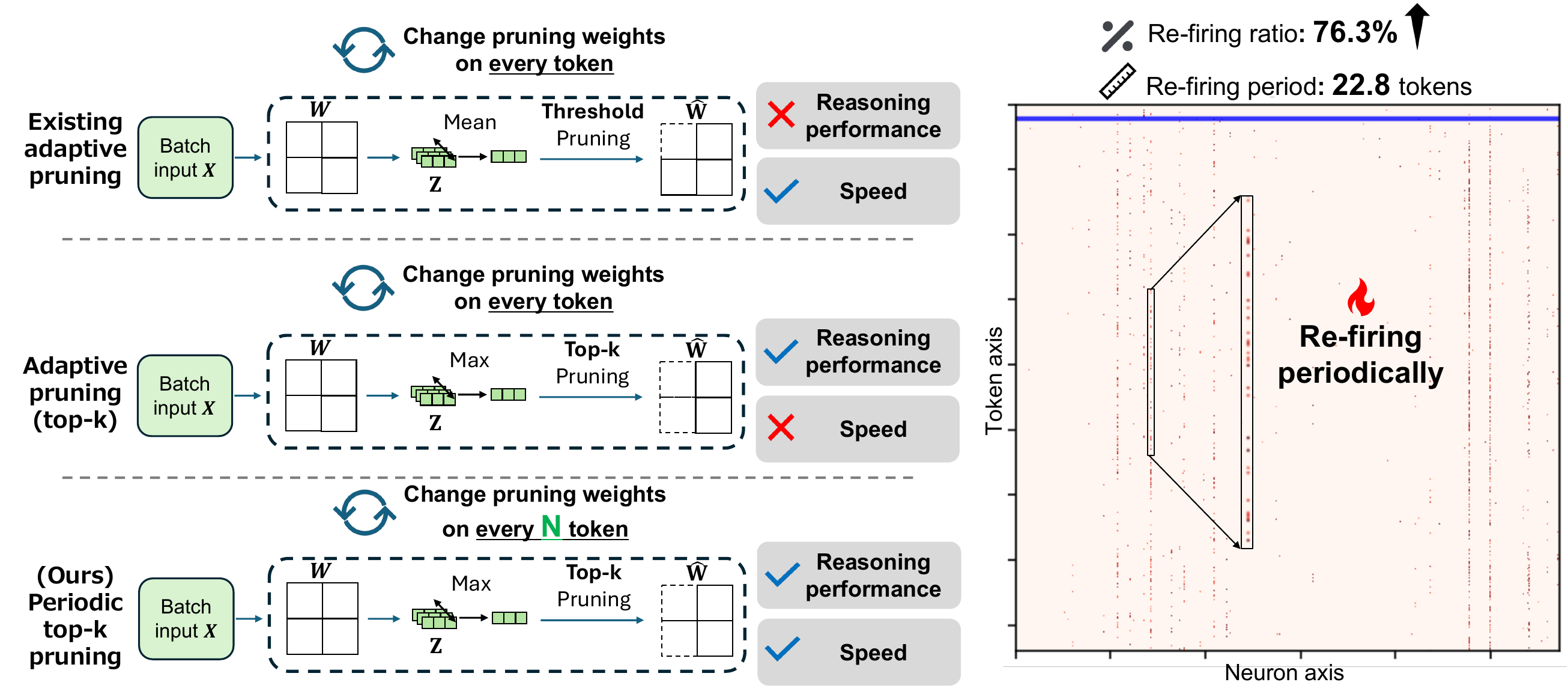}
\caption{\textbf{(Left)} \textit{Top}: existing adaptive pruning~\citep{teal} selects neurons with an offline-calibrated threshold, so the realized sparsity ratio drifts under the aggregated activation distribution and performance degrades. \textit{Middle}: top-$k$ selection is unaffected by that shift, but performing it at every decoding step incurs computational overhead that erases the speedup. \textit{Bottom}: our method updates the pruning mask by top-$k$ selection \emph{periodically} and carries importance across updates through an activation memory, retaining reasoning performance while maintaining computational efficiency. \textbf{(Right)} Activation visualization of $\mathbf{Z}$ in an FFN layer of DeepSeek-R1-Distill-Qwen-7B, where the \textcolor{blue}{blue} horizontal line indicates the input boundary. Neurons with large activation magnitudes (highlighted in \textcolor{red}{red}) exhibit a high re-firing ratio and short re-firing period, motivating our periodic mask update strategy.}
\label{fig:periodic_activation}
\vspace{-6pt}
\end{figure}

\section{Introduction}

Large language models (LLMs) have demonstrated remarkable capabilities across a wide range of tasks~\citep{llama3,qwen2,team2023gemini}, yet they come with substantial computational costs. In particular, recent advances in reasoning capabilities have led to the emergence of Large Reasoning Models (LRMs)~\citep{deepseekr1,openai-o1}, which employ extended chain-of-thought processes to solve complex problems. While effective, these models generate substantially longer outputs, further exacerbating computational costs during inference.
These computational bottlenecks become particularly critical in production settings, where serving multiple requests simultaneously through batched inference (processing multiple input samples together as a single batch) is essential for achieving high serving efficiency~\citep{vllm}.
To further minimize request latency, we need more lightweight models that can run efficiently in batched settings.

 A promising approach to reduce inference cost is pruning.
 Existing pruning methods can be categorized as either \emph{static} or \emph{adaptive}. Static methods~\citep{griffin, wanda, sparsegpt} fix a pruning pattern before decoding, either from calibration data or from the prompt, and reuse it throughout generation; however, we observe that this fixed approach degrades performance on reasoning tasks where activation patterns evolve during chain-of-thought generation. Adaptive methods~\citep{teal,cats} address this by dynamically selecting neurons at each step based on activation magnitudes, achieving strong performance on complex reasoning tasks.

However, existing adaptive pruning methods such as CATS~\citep{cats} and TEAL~\citep{teal} face a fundamental limitation in batched inference scenarios. On GPUs, batched inference requires a single shared pruning mask across all samples in a batch, so per-sample activations must first be aggregated into one score vector.
The failure lies in what happens after aggregation: these methods select neurons with a threshold calibrated offline on an external corpus. That threshold is tuned to a single-sample activation distribution, whereas aggregation across a batch shifts the distribution it is applied to, so the realized sparsity ratio drifts away from the target during generation. Replacing this threshold with top-$k$ selection, while holding every other component of our method fixed, is what allows accuracy to be maintained under batching (Section~\ref{sec:ablation}), identifying threshold-based selection as a primary cause of TEAL's batched collapse.
A second difficulty is specific to reasoning workloads: activation patterns evolve over thousands of generated tokens, so a mask chosen once is stale long before generation ends, while re-selecting it at every decoding step is too expensive to leave any speedup.

In this work, we propose a training-free adaptive pruning method for LRMs that enables efficient batched inference while maintaining strong performance on reasoning tasks. Our contributions are twofold.
(i) We are the first, to our knowledge, to empirically show that existing training-free pruning methods we evaluate severely degrade performance on reasoning tasks under batched inference. The two families fail for different reasons: static methods (Wanda, Griffin; evaluated on the two DeepSeek-R1-Distill models) degrade on reasoning tasks at any batch size, because a pruning pattern fixed before decoding, from calibration data for Wanda and from the prompt for Griffin, cannot track activations that evolve over a long chain of thought, while adaptive methods, which do perform well at batch size 1, collapse once a mask must be shared: TEAL loses 58.4 to 67.7 points of average accuracy relative to the dense model at 50\% target sparsity, across all four models we evaluate.
(ii) We propose a method that retains substantially more accuracy in this regime through two components: \emph{periodic top-$k$ selection}, which is unaffected by the shift that aggregation induces in the activation distribution and runs selection once per update period rather than at every token; and an \emph{activation memory}, which accumulates importance across update phases so that neurons re-activating over time are retained, grounded in the observation that important neurons re-fire periodically during long reasoning generation (Figure~\ref{fig:periodic_activation}; extended visualizations are in Appendix~\ref{sec:activation_visualization}).
Cross-sample aggregation is a third design axis, but not one we claim as a source of our gains: we aggregate with an element-wise maximum rather than a mean (see Batch Aggregation in Section~\ref{sec:importance_score} for details), and our experiments in Section~\ref{sec:ablation} show that accuracy is largely insensitive to this choice. Relying only on runtime activations further makes our method calibration-free, whereas Wanda and TEAL must both calibrate on an external corpus before deployment.


Through extensive experiments on various reasoning tasks, we demonstrate that our method maintains good performance in batched settings, while existing adaptive pruning methods suffer significant performance degradation.
Specifically, our approach outperforms the previous state-of-the-art adaptive pruning method by 39.7 percentage points in average accuracy at batch size 4 with 50\% target sparsity on DeepSeek-R1-Distill-Qwen-7B (DS-R1-Qwen-7B). Additionally, our approach achieves $1.40\times$ speedup over dense inference at 50\% actual sparsity.
The operating regime this targets, and where it does not apply, is stated in Appendix~\ref{sec:limitations}.
\begin{table}[t]
    \centering
    \small
    \caption{Summary of our experiments for various pruning methods across inference settings. Our method ($^{*}$) uniquely maintains consistent performance across all settings, whereas existing approaches degrade on reasoning tasks or batched inference.}
    \label{tab:method_comparison}
    \begin{tabular}{lc|ccc}
        \toprule
        \multirow{2}{*}{\textbf{Task Type}} & \multirow{2}{*}{\textbf{Batch Size}} & \multirow{2}{*}{\shortstack{\textbf{Static}\\\textbf{Pruning}}} & \multirow{2}{*}{\shortstack{\textbf{Existing}\\\textbf{Adaptive Pruning}}} & \multirow{2}{*}{\shortstack{\textbf{  Ours  }}} \\
        & & & & \\
        \midrule
        \multirow{2}{*}{Non-reasoning}
        & 1 & \cmark & \cmark & \cmark \\
        & $\geq$2 & \cmark & \cmark & \cmark \\
        \midrule
        \multirow{2}{*}{Reasoning}
        & 1 & \xmark & \cmark & \cmark \\
        & $\geq$2 & \xmark & \xmark & \cmark$^{*}$ \\
        \bottomrule
    \end{tabular}
\end{table}

\section{Related Work}
\label{sec:related_work}

Pruning methods for large language models can be categorized along several dimensions: granularity, input dependency, and adaptivity. Table~\ref{tab:method_comparison} summarizes key differences between existing approaches and ours.

\paragraph{Pruning Granularity.}
Pruning reduces model size by removing redundant parameters.
\textit{Unstructured pruning} removes individual weights~\citep{sparsegpt, wanda} but requires specialized hardware for acceleration.
\textit{Semi-structured pruning} (e.g., N:M sparsity) also relies on specific GPU architectures~\citep{sparsegpt, wanda}.
\textit{Structured pruning}~\citep{llmpruner, shearedllama} typically requires post-training.
Our method performs structured pruning through standard matrix operations, enabling acceleration on any hardware without specialized kernels or post-training.

\paragraph{Input-Independent vs. Input-Dependent Pruning.}
\textit{Input-independent} methods determine a fixed pruning pattern before inference using calibration data~\citep{wanda, slicegpt, shortenedllama, sparsegpt}, benefiting from batch compatibility but unable to exploit input-specific activation patterns.
\textit{Input-dependent} methods dynamically select neurons based on runtime activations~\citep{griffin, teal, cats, prosparse}, but introduce challenges for batched inference, as different samples require distinct pruning patterns.

\paragraph{Static vs. Adaptive Pruning.}
\textit{Static pruning} determines a fixed sparsity pattern and reuses it throughout generation~\citep{griffin}, while \textit{adaptive pruning} dynamically updates the mask during decoding~\citep{cats, teal, dejavu}.
Static pruning cannot adapt to evolving activation patterns, leading to degradation on reasoning models with long outputs.
Existing adaptive methods rely on threshold-based selection, which suffers from distribution shifts when aggregating across batches.
Moreover, DejaVu~\citep{dejavu} and ProSparse~\citep{prosparse} operate at the individual sample level, making batched inference challenging.
Our method addresses these limitations through periodic top-$k$ selection combined with an activation memory that carries importance across mask updates, as detailed in Section~\ref{sec:method}.

\begin{figure}[t]
    \centering
    \includegraphics[width=0.9\textwidth]{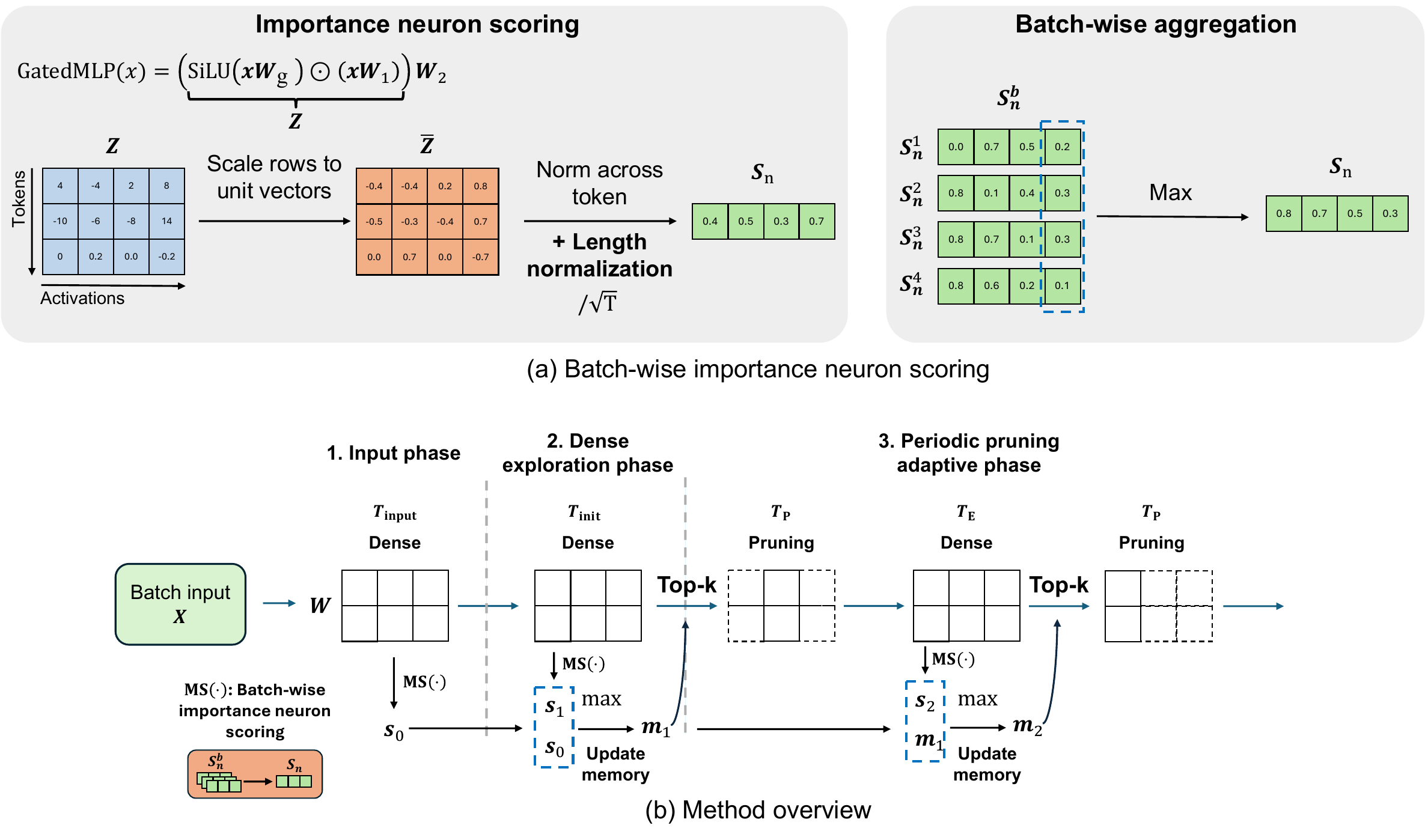}
    \caption{Overview of our batch-wise adaptive pruning approach. (a) Importance scores are computed per sample and aggregated into a single score vector, since a batch must share one pruning mask; we use an element-wise maximum as the default aggregation operator. (b) The method operates in three phases: input processing, dense exploration, and periodic adaptive pruning, with an activation memory carrying importance scores across successive mask updates.}
    \label{fig:method_overview}
\end{figure}

\section{Method}
\label{sec:method}

In this section, we present our batch-wise adaptive pruning approach for efficient LLM inference. Our method is training-free, requiring no fine-tuning; pruning decisions are made in real time, relying solely on model activations observed during inference.
We first introduce the preliminaries on gated MLP structures (Section~\ref{sec:preliminaries}), which account for the majority of the parameters in modern Transformer-based LLMs and thus are the pruning target of our method.
Next, we describe our batch-wise importance scoring mechanism (Section~\ref{sec:importance_score}), which defines the metric to determine which neurons should be pruned in the gated MLP for batched inference.
Finally, we detail our three-phase pruning algorithm (Section~\ref{sec:algorithm}), which employs periodic pruning using the defined batch-wise importance scores.

\subsection{Preliminaries}
\label{sec:preliminaries}

Modern transformer architectures employ gated MLP structures in their feedforward blocks~\citep{llama3, qwen2}. Given a sequence of input hidden states $\mathbf{X} \in \mathbb{R}^{T \times D}$ where $T$ is the number of tokens, each hidden state $\mathbf{x} \in \mathbb{R}^{D}$ is processed by the gated MLP:
\begin{align}
    \mathbf{z} = \sigma(\mathbf{W}_g \mathbf{x}) \odot (\mathbf{W}_1 \mathbf{x})
\end{align}
where $\sigma$ denotes the SiLU activation function and $\odot$ signifies element-wise multiplication. For all weight matrices, $\mathbf{W}_1, \mathbf{W}_g \in \mathbb{R}^{D_{\text{FF}} \times D}$ and $\mathbf{W}_2 \in \mathbb{R}^{D \times D_{\text{FF}}}$ where typically $D_{\text{FF}} \gg D$. We refer to $\mathbf{z} = \text{FF}_1(\mathbf{x}) \in \mathbb{R}^{D_{\text{FF}}}$ as the FF activations, and the output is computed as $\text{FF}_2(\mathbf{z}) = \mathbf{W}_2 \mathbf{z}$.

\subsection{Batch-wise Importance Neuron Scoring}
\label{sec:importance_score}

We define an \emph{importance score} $\mathbf{s} \in \mathbb{R}^{D_{\text{FF}}}$ as a measure of how critical each neuron in the FF layer is for the current task. Neurons with low importance scores are pruned to reduce computational cost while retaining the most task-relevant neurons. As illustrated in Figure~\ref{fig:method_overview}(a), following prior work on activation-based pruning~\citep{griffin}, we compute importance scores by aggregating normalized activation magnitudes across tokens.

For a sequence of $T$ tokens, let $\mathbf{Z} = \text{FF}_1(\mathbf{X}) \in \mathbb{R}^{T \times D_{\text{FF}}}$ denote the FF activations for the full sequence. We first compute the relative activations $\overline{\mathbf{Z}}$ by applying row-wise $\ell_2$ normalization: $[\overline{\mathbf{Z}}]_t = [\mathbf{Z}]_t / \|[\mathbf{Z}]_t\|_2$ for each token $t$. This normalization ensures that we capture the relative importance of each neuron within a token, rather than being dominated by tokens with large overall activation magnitudes. As described in Section~\ref{sec:algorithm}, our method operates in three distinct phases, and we compute importance scores at each phase separately. The importance score at each phase $n$ is then computed by taking the column-wise $\ell_2$-norm:

\begin{equation}
    \mathbf{s}_n = \text{MS}(\mathbf{Z}) = \frac{1}{\sqrt{T}} \left[ \|[\overline{\mathbf{Z}}]_{\cdot, 1}\|_2, \ldots, \|[\overline{\mathbf{Z}}]_{\cdot, D_{\text{FF}}}\|_2 \right]^\top
    \label{eq:importance_score}
\end{equation}

where $n$ denotes the phase index, and $\text{MS}(\cdot)$ denotes the batch-wise scoring function. Intuitively, neurons with consistently high relative activations across multiple tokens receive higher importance scores, capturing ``persistently important'' neurons that contribute meaningfully to the model's computation. Unlike prior work~\citep{griffin} that computes importance scores over the entire sequence in a single pass, our method computes importance scores at each phase separately. Since different phases may have varying numbers of tokens, we normalize by $\sqrt{T}$ (the square root of the number of tokens) to eliminate length bias, enabling fair comparison of importance scores across phases.

\paragraph{Batch Aggregation.}
Because all samples in a batch must share one pruning mask, the per-sample importance scores have to be reduced to a single vector before selection; the choice of reduction operator is therefore a required design decision rather than an optional addition. We adopt the element-wise maximum as our default, as it is the conservative choice under a shared mask: a neuron scored highly by a single sample keeps that score after aggregation instead of being diluted by the remaining $B-1$ samples, which is intended to reduce the chance that a neuron critical to one sample is dropped because the others do not use it. Selection is still by top-$k$, so such a neuron survives only if its aggregated score is among the top $k$. Which operator is preferable is task-dependent: max and mean each lead on different benchmarks and on different models, and their cross-model averages differ by 1.2 points (Section~\ref{sec:ablation}), so the aggregation operator is not the dominant factor in our method's performance. Unless stated otherwise, all experiments in this paper use the maximum. The components responsible for the gains are the periodic top-$k$ selection and the activation memory introduced below.

For batched inference with $B$ samples, we aggregate importance scores across the batch, excluding padding tokens and EOS tokens from the aggregation. For a single sample ($B\!=\!1$), we directly use $\mathbf{s}_n$. For multiple samples ($B\!>\!1$), we aggregate using element-wise maximum:
\begin{align}
    \bar{\mathbf{s}}_n = \max_{b \in \{1, \ldots, B\}} \mathbf{s}_n^{(b)}
\end{align}
where $\mathbf{s}_n^{(b)}$ denotes the importance score for sample $b$ at phase $n$. We then construct a \emph{pruning mask} $\mathbf{M} \in \{0, 1\}^{D_{\text{FF}}}$, a binary vector indicating which neurons to retain. Specifically, we select the top-$k$ neurons with the highest importance scores, where $k = \lfloor (1 - \rho) \cdot D_{\text{FF}} \rfloor$ and $\rho \in (0, 1)$ is the target sparsity ratio. The shared pruning mask is then applied uniformly across all samples in the batch. Top-$k$ selection reads only the relative ordering of the aggregated scores, not their absolute scale, so it is unaffected when aggregation shifts the activation distribution. A threshold calibrated on a different distribution is not, which is what threshold-based selection loses under batching; Section~\ref{sec:ablation} isolates the effect.

\subsection{Batch-wise Adaptive Pruning Algorithm}
\label{sec:algorithm}

Our batch-wise adaptive approach operates in three distinct phases during autoregressive generation, as illustrated in Figure~\ref{fig:method_overview}(b). The key insight is that we balance computational efficiency with adaptivity by alternating between sparse and dense exploration phases.

\paragraph{Phase 1: Input Phase.}
During the initial prompt processing with $T_{\text{input}}$ input tokens, we compute the FF activations $\mathbf{z}_t$ for each token $t = 1, \ldots, T_{\text{input}}$. Using these activations, we compute the initial importance score $\mathbf{s}_0$ using the batch-wise scoring function (Equation~\ref{eq:importance_score}).
\paragraph{Phase 2: Dense Exploration Phase.}
After the prompt, we perform $T_{\text{init}}$ steps of dense computation. This dense exploration phase serves two purposes: (1) it allows the model to generate initial tokens using full capacity, which is particularly important for establishing the direction of reasoning, and (2) it enables us to collect activation statistics that better reflect the generation context rather than just the input. At the end of this phase, we compute a new importance score $\mathbf{s}_1$ from the collected activations and update our \emph{activation memory} $\mathbf{m} \in \mathbb{R}^{D_{\text{FF}}}$ using an element-wise maximum: $\mathbf{m}_1 \leftarrow \max(\mathbf{s}_0, \mathbf{s}_1)$. The activation memory $\mathbf{m}$ serves as a buffer that accumulates neuron importance information across phases, ensuring that neurons identified as important in earlier phases remain candidates for selection while incorporating new information from subsequent phases. We then generate the initial pruning mask $\mathbf{M}_1 = \text{top-}k(\mathbf{m}_1, k)$ and immediately begin sparse computation.

\paragraph{Phase 3: Periodic Adaptive Pruning Phase.}
As shown in Figure~\ref{fig:periodic_activation}, neurons with high activation magnitudes exhibit periodic patterns during autoregressive generation. Based on this observation, we periodically update the pruning mask to adapt to these evolving patterns, enabling effective performance even for reasoning tasks that generate long outputs.

After the initial dense exploration, we alternate between cycles of sparse and dense computation. Each cycle $n$ consists of two stages: (1) \emph{pruning}, where we perform $T_p$ steps of sparse forward passes using the current pruning mask, and (2) \emph{exploration}, where we perform $T_E$ steps of dense computation while collecting activations to update the importance scores. At the end of each exploration stage, we update the activation memory using the maximum of the current score and the previous memory: $\mathbf{m}_n \leftarrow \max(\mathbf{m}_{n-1}, \mathbf{s}_n)$, and generate a new pruning mask $\mathbf{M}_n = \text{top-}k(\mathbf{m}_n, k)$. This update rule accumulates importance information across multiple cycles, ensuring that persistently important neurons are retained while allowing the mask to adapt to evolving activation patterns. Specifically, the pruning mask is updated every $T_{\text{trans}} = T_E + T_p$ steps, alternating between sparse pruning and dense exploration phases. We set $T_{\text{trans}}=20$ based on the empirically observed median firing period of important neurons; a detailed analysis is provided in Section~\ref{sec:results}.

During the pruning stage, we leverage the pruning mask to reduce computational costs. Given a mask $\mathbf{M}$, we define the set of retained neuron indices as $\mathcal{I} = \{i : \mathbf{M}[i] = 1\}$, where $|\mathcal{I}| = (1 - \rho) \cdot D_{\text{FF}}$. The pruned weight matrices are obtained by selecting the corresponding rows: $\widehat{\mathbf{W}}_g, \widehat{\mathbf{W}}_1 \in \mathbb{R}^{k \times D}$ and $\widehat{\mathbf{W}}_2 \in \mathbb{R}^{D \times k}$. The sparse forward pass then computes:
\begin{align}
    \widehat{\mathbf{z}} = \sigma(\widehat{\mathbf{W}}_g \mathbf{x}) \odot (\widehat{\mathbf{W}}_1 \mathbf{x}), \quad \mathbf{y} = \widehat{\mathbf{W}}_2 \widehat{\mathbf{z}}
\end{align}
where $\widehat{\mathbf{z}} \in \mathbb{R}^{k}$. This structured pruning approach enables efficient matrix operations on modern GPUs, as the reduced dimensions lead to proportionally smaller computations.


\section{Experiment}
\label{sec:Experiment}

\subsection{Models and Datasets}

We evaluate our proposed batch-wise adaptive pruning method on two reasoning models: DeepSeek-R1-Distill-Qwen-7B (DS-R1-Qwen-7B) and DeepSeek-R1-Distill-Llama-8B (DS-R1-Llama-8B)~\citep{deepseekr1}. We additionally evaluate on Qwen3-1.7B and Qwen3-8B~\citep{qwen3}, with results reported in Appendix~\ref{sec:qwen3_results_pruning}.

For evaluation, we use diverse reasoning benchmarks following the prior work evaluation framework~\citep{passk}. For mathematical reasoning, we use TinyGSM8K~\citep{gsm8k,tinybenchmarks}, a 100-sample subset of GSM8K (grade school math word problems) selected to reduce computational costs while maintaining evaluation reliability. We also include MATH500~\citep{math} for competition-level mathematics, MINERVA Math~\citep{minerva} for scientific reasoning, and AMC23~\citep{math} from the 2023 American Mathematics Competition. For general reasoning, we use GPQA-DIAMOND~\citep{gpqa}, a graduate-level science QA benchmark, following Open R1~\citep{openr1} evaluation settings.


\begin{figure}[t]
    \centering
    \includegraphics[width=0.9\linewidth]{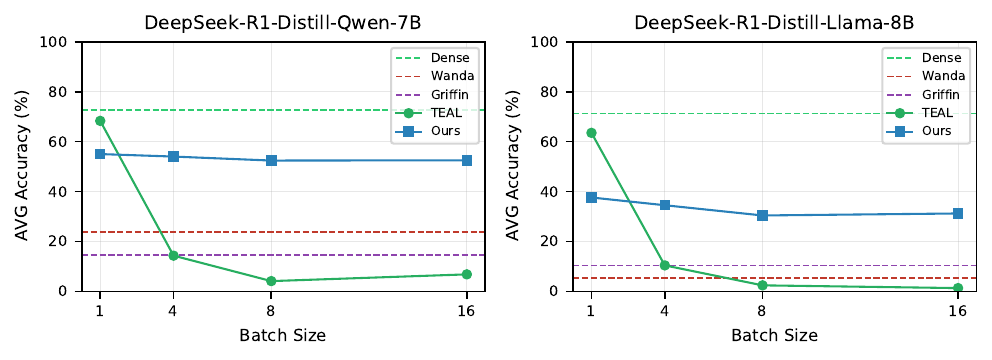}
    \caption{Average accuracy across the five reasoning tasks with varying batch sizes at 50\% target sparsity. (Left) DeepSeek-R1-Distill-Qwen-7B, (Right) DeepSeek-R1-Distill-Llama-8B.}
    \label{fig:accuracy-batchsize}
\end{figure}

\subsection{Baselines}

Because our proposed method is training-free, we compare against training-free pruning methods. Wanda~\citep{wanda} is an input-independent method that applies 2:4 structured sparsity after calibrating on C4. Griffin~\citep{griffin} employs input-dependent but static pruning, capturing important neurons during the prompt phase and reusing the pruned weights throughout decoding. TEAL~\citep{teal} is an adaptive method that uses threshold-based selection, calibrating on C4 to extract activation thresholds and pruning neurons below the threshold during inference with uniform sparsity across all layers. For batched evaluation, following~\citep{teal}, we aggregate activations within a batch using mean, excluding padding and EOS tokens, then apply the single-sample algorithm. Since Dense and Wanda performances are batch-size independent, we report their results using batch size 4.

\paragraph{Target sparsity versus actual sparsity.}
Two notions of sparsity appear throughout this paper and induce different speedups, so we define both. \emph{Target sparsity} is the ratio specified before inference: for our method it sets how many neurons the top-$k$ mask removes, for TEAL the C4-calibrated activation threshold. \emph{Actual sparsity} is the fraction of parameters effectively pruned over the whole generation; it falls below the target for our method, which alternates dense and sparse steps and prunes only the FFN block, and drifts from it for TEAL whenever the runtime distribution departs from calibration. Speedups measured under the two settings are therefore not directly comparable. Section~\ref{sec:results} reports throughput at matched \emph{actual} sparsity; throughput at 50\% \emph{target} sparsity is given in Appendix~\ref{sec:target_sparsity_speed}. We label which setting is used in every figure and table.

\subsection{Results}
\label{sec:results}

\begin{table}[tb]
    \centering
    \caption{Pruning performance comparison at 50\% target sparsity with batch size 4. Our method significantly outperforms all pruning baselines (Wanda, Griffin, TEAL) across multiple reasoning benchmarks on both DeepSeek-R1-Distill-Qwen-7B and DeepSeek-R1-Distill-Llama-8B models. Bold values indicate the highest performance among all pruning methods.}
    \label{tab:main-result}
    \resizebox{\columnwidth}{!}{%
    \begin{tabular}{l|l|ccccc|c}
    \toprule
    \multirow{2}{*}{\textbf{Model}} & \multirow{2}{*}{\textbf{Method}} & \multicolumn{5}{c|}{\textbf{Tasks}} & \multirow{2}{*}{\textbf{AVG}} \\
    \cmidrule(lr){3-7}
    & & \textbf{GSM8K} & \textbf{MATH500} & \textbf{MINERVA} & \textbf{AMC23} & \textbf{GPQA-DIAMOND} & \\
    \midrule

    \multirow{5}{*}{\textit{\makecell[l]{DeepSeek-\\R1-Distill-\\Qwen-7B}}}
    & Dense & 92.0 & 91.8 & 39.7 & 87.5 & 52.5 & 72.7 \\
    & Wanda & 59.0 & 24.4 & 7.4 & 12.5 & 15.7 & 23.8 \\
    & Griffin & 22.0 & 14.8 & 11.0 & 10.0 & 15.2 & 14.6 \\
    & TEAL & 30.0 & 11.2 & 4.8 & 10.0 & 15.7 & 14.3 \\
    & Ours & \textbf{89.0} & \textbf{71.0} & \textbf{29.4} & \textbf{50.0} & \textbf{30.8} & \textbf{54.0} \\
    \midrule

    \multirow{5}{*}{\textit{\makecell[l]{DeepSeek-\\R1-Distill-\\Llama-8B}}}
    & Dense & 96.0 & 91.0 & 33.8 & 90.0 & 45.5 & 71.3 \\
    & Wanda & 9.0 & 4.8 & 0.7 & 5.0 & 7.1 & 5.3 \\
    & Griffin & 16.0 & 11.8 & 4.8 & 2.5 & 16.7 & 10.4 \\
    & TEAL & 28.0 & 5.4 & 2.2 & 0.0 & 16.2 & 10.4 \\
    & Ours & \textbf{75.0} & \textbf{38.8} & \textbf{12.1} & \textbf{25.0} & \textbf{21.7} & \textbf{34.5} \\
    \bottomrule
    \end{tabular}%
    }
    \end{table}

\paragraph{Activation periodicity and re-firing ratio.}
Our method updates the pruning mask periodically based on the observation that important neurons tend to re-fire over time. We quantify this by measuring (1) the re-firing ratio, the proportion of important neurons (with relative activation $\bar{\mathbf{Z}} > 0.05$) that activate at least twice during generation, and (2) their median firing period, using 10 samples per benchmark.

\begin{wraptable}[10]{r}{0.49\textwidth}
    \centering
    \caption{Median firing period and re-firing ratio of important neurons ($\bar{\mathbf{Z}} > 0.05$). Mean $\pm$ std over 10 samples $\times$ 5 benchmarks.}
    \label{tab:periodicity}
    \small
    \resizebox{\linewidth}{!}{%
    \begin{tabular}{l|cc}
    \toprule
    \textbf{Model} & \textbf{Median Period} & \textbf{Re-firing (\%)} \\
    \midrule
    \makecell[l]{DeepSeek-R1\\Distill-Qwen-7B}  & 22.8 $\pm$ 1.9 & 76.3 $\pm$ 2.8 \\
    \midrule
    \makecell[l]{DeepSeek-R1\\Distill-Llama-8B} & 20.8 $\pm$ 2.9 & 71.9 $\pm$ 2.2 \\
    \bottomrule
    \end{tabular}}
\end{wraptable}

As shown in Table~\ref{tab:periodicity}, the re-firing ratio exceeds 71\% across both models, with a median firing period of 20--23 tokens. We thus set $T_{\text{trans}}=20$, slightly below the observed median, to refresh the mask before important neurons miss their re-firing window. Extended results are in Appendix~\ref{sec:qwen3_periodicity}.

\paragraph{Our method outperforms baselines in batched settings.}
Table~\ref{tab:main-result} presents the performance comparison at 50\% target sparsity with batch size 4. Our proposed method significantly outperforms all baseline approaches across multiple reasoning benchmarks on both DS-R1-Qwen-7B and DS-R1-Llama-8B. On DS-R1-Qwen-7B, our method achieves an average accuracy of 54.0\% across the five tasks, compared to TEAL's 14.3\%, an improvement of 39.7 points. Similarly, on DS-R1-Llama-8B, our approach attains 34.5\% average accuracy versus TEAL's 10.4\%, an improvement of 24.1 points. We additionally evaluate on non-reasoning benchmarks in Appendix~\ref{sec:nonreasoning}, confirming that our method maintains performance within approximately 1\% of the dense baseline at 50\% sparsity on non-reasoning tasks. Static pruning methods such as Wanda and Griffin exhibit poor performance on reasoning tasks, as they cannot adapt to the evolving activation patterns during long chain-of-thought generation. Griffin achieves only 14.6\% and 10.4\% average accuracy on DS-R1-Qwen-7B and DS-R1-Llama-8B respectively, while Wanda obtains 23.8\% and 5.3\%. Although TEAL employs adaptive threshold-based pruning, its performance degrades significantly in batched settings due to distribution shifts when aggregating activations across multiple samples. We further validate our method on Qwen3-1.7B and Qwen3-8B, where our method outperforms TEAL by 41.8 and 45.5 points on average, respectively; detailed results are provided in Appendix~\ref{sec:qwen3_results_pruning}.

The difference is qualitative as well as numerical: at this operating point TEAL's generations collapse into degenerate repetition loops on 96--99\% of MATH500 samples across all four models, whereas ours do so on 2.2--7.4\% and otherwise maintain coherent chain-of-thought (Appendix~\ref{sec:trajectory_quality}).

\begin{figure}[tb]
    \centering
    \includegraphics[width=0.85\linewidth]{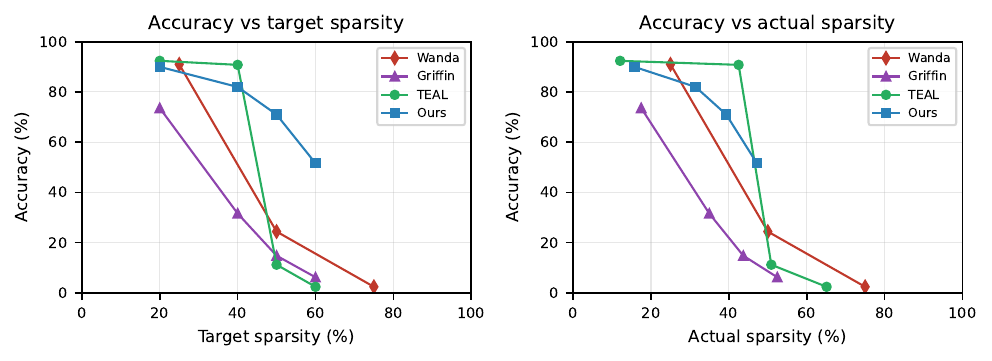}
    \caption{Accuracy on MATH500 across sparsity ratios with batch size 4 on DeepSeek-R1-Distill-Qwen-7B. (Left) Accuracy vs.\ target sparsity, (Right) Accuracy vs.\ actual sparsity. Target sparsity refers to the intended pruning ratio set before inference, while actual sparsity indicates the proportion of parameters effectively pruned during generation.}
    \label{fig:sparsity-ratio}
\end{figure}

\paragraph{Our method maintains consistent performance across all batch sizes.}
Figure~\ref{fig:accuracy-batchsize} illustrates how average accuracy varies with batch size at 50\% target sparsity. TEAL demonstrates high performance at batch size 1 but experiences substantial degradation as batch size increases, due to the distribution shift when aggregated activations diverge from single-sample calibration settings. In contrast, our method maintains robust and consistent performance across all batch sizes, demonstrating its practical advantage for batched deployment. Detailed results are in Appendix Table~\ref{tab:full-result}.

\paragraph{Our method maintains strong performance up to high sparsity levels.}
Figure~\ref{fig:sparsity-ratio} shows performance on MATH500 across different target sparsity levels with batch size 4 on DS-R1-Qwen-7B. TEAL and other baselines maintain reasonable performance at low sparsity levels, for example TEAL achieves 92.4\% and 90.8\% at 20\% and 40\% target sparsity, respectively. However, they suffer from sharp performance degradation beyond 40\% target sparsity; TEAL drops to 11.2\% at 50\% and 2.4\% at 60\%. In contrast, our method sustains strong performance even at high sparsity, achieving 71.0\% accuracy at 50\% and 51.6\% at 60\% target sparsity, exhibiting a significantly more gradual and stable accuracy decline as sparsity increases compared to all baselines. At 20\% target sparsity our method is within 1.8 points of the dense baseline, so the gap narrows as target sparsity is reduced rather than reflecting a fixed accuracy ceiling.

\begin{figure}[b]
    \centering
    \includegraphics[width=0.85\linewidth]{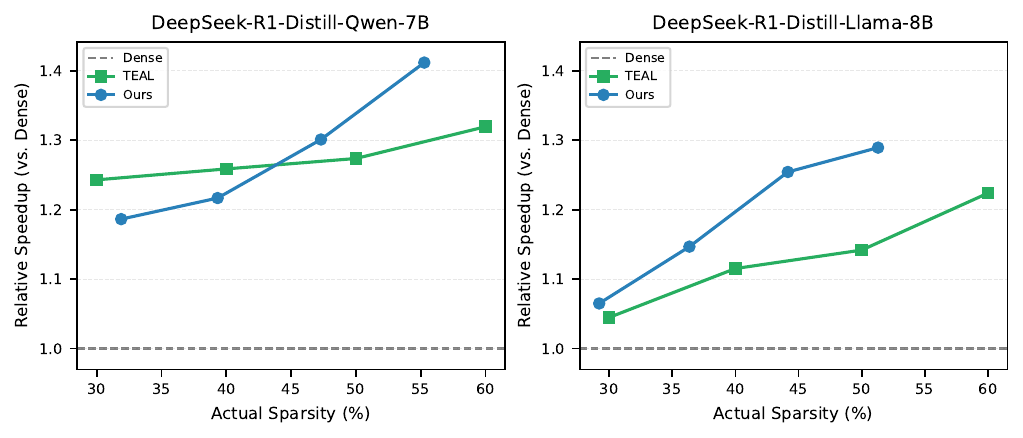}
    \caption{Generation throughput speedup vs.\ actual sparsity at batch size 4. (Left) DeepSeek-R1-Distill-Qwen-7B, (Right) DeepSeek-R1-Distill-Llama-8B.}
    \label{fig:speed-actual-sparsity}
\end{figure}

\paragraph{Our method achieves practical speedup in batched scenarios.}
We evaluate end-to-end decoding speed on an NVIDIA H100 GPU following the GPT-Fast~\citep{gptfast} benchmarking setup.
Figure~\ref{fig:speed-actual-sparsity} shows throughput speedup versus actual sparsity at batch size 4. Except at low sparsity on DS-R1-Qwen-7B, our method achieves higher speedup than TEAL across sparsity levels on both models. At approximately 50\% actual sparsity, our method achieves $1.29\times$ speedup on DS-R1-Llama-8B and $1.40\times$ on DS-R1-Qwen-7B. As batch size increases and the workload shifts from memory-bound to compute-bound, reducing memory loads by skipping near-zero activations (as in TEAL) becomes less effective, while structurally reducing parameters via smaller weight matrices (as in our method) directly reduces computation, giving our method a practical efficiency advantage in batched deployment. Furthermore, this advantage grows with both batch size and sparsity level; detailed results across batch sizes are provided in Appendix~\ref{sec:actual_sparsity_speed_all_bs}. The computational overhead of our adaptive mask computation is negligible ($<$0.002\% of dense MLP FLOPs; see Appendix~\ref{sec:flops_overhead}). Detailed throughput values at 50\% target sparsity and speed-accuracy trade-offs are provided in Appendix~\ref{sec:target_sparsity_speed}.
\subsection{Ablation Study}
\label{sec:ablation}

\paragraph{Batch aggregation method.}
Top of Table~\ref{tab:design_ablation} compares batch-wise aggregation strategies at 50\% target sparsity with batch size 4.
The two are within 1.2 points in cross-model average and the per-model winner is split: mean is better on DS-R1-Qwen-7B (54.8 versus 54.0), max on DS-R1-Llama-8B (34.5 versus 31.4). Per-task differences are similarly mixed, with max ahead on GSM8K (+8.5) and AMC23 (+2.5) and mean ahead on MATH500 (+3.2) and GPQA (+1.5) points (Appendix Table~\ref{tab:design_ablation_full}). We report max as the default because some reduction is unavoidable under a shared mask and max is the conservative choice; selecting the operator adaptively is left to future work.

\begin{table}[t]
    \centering
    \caption{Performance comparison of aggregation methods (mean/max) and activation memory (on/off) at 50\% target sparsity with batch size 4. Values are averaged over 5 benchmarks. Bold indicates best per model.}
    \label{tab:design_ablation}
    \small
    \begin{tabular}{l|c|c}
    \toprule
    Setting & \makecell{DeepSeek-R1\\Distill-Qwen-7B} & \makecell{DeepSeek-R1\\Distill-Llama-8B} \\
    \midrule
    Max (Ours) & 54.0 & \textbf{34.5} \\
    Mean       & \textbf{54.8} & 31.4 \\
    \midrule
    Memory (Ours) & \textbf{54.0} & \textbf{34.5} \\
    No Memory     & 45.3 & 25.5 \\
    \bottomrule
    \end{tabular}
\end{table}

\paragraph{Activation memory.}
Bottom of Table~\ref{tab:design_ablation} evaluates the effect of activation memory by comparing the proposed memory-based approach with a memory-free variant, where the pruning mask is reconstructed independently at every transition step. Enabling activation memory improves average accuracy by 9.0 points for DS-R1-Llama-8B and 8.7 points for DS-R1-Qwen-7B. This result indicates that memory stabilizes pruning decisions and improves robustness on reasoning tasks by preserving neurons that re-activate over time.

\paragraph{Hyperparameter and threshold ablation.}
We conduct hyperparameter ablation on $T_{\text{init}}$, $T_E$, and $T_{\text{trans}}$, measuring accuracy and speed under each variation. Notably, on DS-R1-Qwen-7B with MATH500, our periodic configuration achieves accuracy within 17.2 points of the most frequent update setting while being significantly faster than the dense baseline, confirming that periodic updates effectively balance accuracy and speed; detailed results are provided in Appendix~\ref{sec:hyperparameter_ablation}.
We also evaluate threshold-based pruning calibrated on C4 following TEAL's approach. Since the threshold is calibrated at batch size 1, the activation distribution shifts after batch aggregation, resulting in accuracy of at most 3.2\% on every benchmark at 50\% target sparsity. Detailed results are provided in Appendix~\ref{sec:threshold_ablation}.

\section{Conclusion}

In this work, we presented a training-free batch-wise adaptive pruning method for batched inference in LRMs. We first showed that the training-free pruning baselines we evaluate all degrade severely on reasoning tasks under batched inference: static methods degrade at any batch size, while adaptive methods work at batch size 1 and collapse once a mask is shared, with TEAL losing 58.4 to 67.7 points of average accuracy relative to the dense model at 50\% target sparsity. We traced the latter collapse to threshold-based selection applied after cross-sample aggregation. We then proposed a method built on periodic top-$k$ selection, which is unaffected by the shift that aggregation induces in the activation distribution while running selection only once per update period, and an activation memory that retains neurons re-firing across update phases.

Our experiments show that this design retains substantially more accuracy than prior methods in the high-sparsity batched regime, and that its speedup keeps growing with sparsity where activation-skipping methods saturate. We view the contribution as opening batch-wise adaptive pruning as a problem and providing a first working method in it.


\bibliography{adaptive_pruning}
\bibliographystyle{colm2026_conference}

\appendix
\newpage

\section{Limitations}
\label{sec:limitations}

Our proposed method achieves efficient inference for reasoning models and tasks in batched settings by periodically pruning weights based on activation patterns observed during generation. However, some limitations remain.

This approach requires direct access to intermediate activations within the model, which limits its applicability to closed-source or API-based models where internal states are not exposed.

In the low-sparsity regime our method is not the best option. Below roughly 40\% target sparsity TEAL remains ahead of our method in accuracy on MATH500 (Figure~\ref{fig:sparsity-ratio}), and on DS-R1-Qwen-7B its activation skipping is also faster there, since our periodic updates and top-$k$ selection add overhead that a low sparsity level does not yet justify. TEAL is also ahead at batch size 1 (Table~\ref{tab:full-result}), where a per-token mask update is essentially free for its threshold but not for our top-$k$ selection, which we therefore run only once per update period. Pruning is typically deployed at high sparsity, which is the regime this paper targets and where the ordering reverses: beyond 40\% target sparsity TEAL collapses on reasoning tasks while our method retains accuracy.

A gap to the dense model also still remains: at 50\% target sparsity with batch size 4, average accuracy over the five reasoning benchmarks is 54.0 for our method against 72.7 for the dense model on DS-R1-Qwen-7B, and 34.5 against 71.3 on DS-R1-Llama-8B (Table~\ref{tab:main-result}). Maintaining accuracy under batched inference at high sparsity on reasoning tasks is intrinsically difficult, and we read this gap as reflecting the difficulty of the setting rather than a shortcoming specific to our method; it is also smaller than the gap of any prior method evaluated under the same conditions. The gap is also concentrated on DS-R1-Llama-8B: on the other three models we evaluate, our method stays within 19 to 23 points of dense, whereas TEAL loses 58 to 68 points on all four (Table~\ref{tab:qwen3_pruning}). Closing the remaining distance to dense accuracy is an explicit direction for follow-up work.

\section{Reasoning Trajectory Quality}
\label{sec:trajectory_quality}

Average accuracy reports whether a final answer is correct, but not how the generated reasoning fails when it is wrong. To characterize the failure modes, we measure the rate of degenerate repetition on MATH500 generations at 50\% target sparsity with batch size 4. A generation is counted as degenerate if the ratio of repeated word 4-grams to total 4-grams is at least 0.8, which detects the case where decoding enters a loop and emits the same fragment until the length limit.

\begin{table}[h]
\centering
\caption{Rate of degenerate-repetition collapse (\% of generations with word 4-gram repetition ratio $\geq 0.8$; lower is better) on MATH500 at 50\% target sparsity with batch size 4.}
\label{tab:repetition}
\small
\begin{tabular}{l|cc}
\toprule
\textbf{Model} & \textbf{Ours} & \textbf{TEAL} \\
\midrule
DeepSeek-R1-Distill-Qwen-7B  & \textbf{2.2} & 96.8 \\
DeepSeek-R1-Distill-Llama-8B & \textbf{3.8} & 99.2 \\
Qwen3-1.7B                   & \textbf{7.4} & 98.8 \\
Qwen3-8B                     & \textbf{4.6} & 96.4 \\
\bottomrule
\end{tabular}
\end{table}

Table~\ref{tab:repetition} shows that the two methods fail in qualitatively different ways at the same operating point. Under TEAL, essentially every generation degenerates (96.4--99.2\% across the four models), whereas under our method the rate stays below 10\%. The typical TEAL failure is a coherent prefix that diverges into a short fragment repeated until the maximum generation length is reached (for example, the fragment ``Wait, no, I.'' continued for tens of thousands of characters), rather than a plausible chain of thought that arrives at a wrong answer. This indicates that the accuracy collapse reported in Section~\ref{sec:results} understates the difference between the two settings: TEAL's remaining accuracy at this sparsity does not correspond to degraded but usable reasoning, while our method continues to produce coherent chain-of-thought.

Table~\ref{tab:doomloop} illustrates this on a single MATH500 problem, comparing the end of each generation at 50\% target sparsity with batch size 4 on DS-R1-Qwen-7B. TEAL reaches the correct value but never emits it as an answer: the generation enters a loop that repeats the same fragment until the length limit, ending after 28{,}560 characters and being graded incorrect. Our method closes the same derivation in 13{,}103 characters.

\begin{table}[!ht]
\centering
\caption{End of the generation for MATH500 problem 75 (level 4), ``Two fair, 6-sided dice are thrown. What is the probability that the product of the two numbers is a multiple of 5?'', whose answer is 11/36, at 50\% target sparsity with batch size 4 on DS-R1-Qwen-7B.}
\label{tab:doomloop}
\small
\begin{tabular}{@{}p{0.1\textwidth}p{0.82\textwidth}@{}}
\toprule
\textbf{Method} & \textbf{Last characters of the generation} \\
\midrule
TEAL & \texttt{\ldots{} Wait, 11/36. Wait, 11/36. Wait, 11/36. Wait, 11/36. Wait, 11/36. Wait, 11/36. Wait, 11/36. Wait, 11/36. Wait, 11} \newline (cut off at the generation length limit; graded incorrect) \\
\midrule
Ours & \texttt{\ldots{} we subtract this from 1: 1 - 25/36 = 36/36 - 25/36 = 11/36. Thus, the probability that the product of the two numbers is a multiple of 5 is \textbackslash boxed\{11/36\}.} \\
\bottomrule
\end{tabular}
\end{table}

\section{Composition with KV-Cache Compression}
\label{sec:rkv_composition}

R-KV~\citep{rkv} addresses a different objective from ours: it reduces inference latency by compressing the KV cache, which is effective when KV-cache memory is the dominant bottleneck, typically in long-context or memory-constrained serving. In settings where KV-cache memory is not a bottleneck, R-KV provides no speedup. Our method, in contrast, prunes neurons in the model weights themselves, so it accelerates inference regardless of the serving environment.

Because the two prune along orthogonal axes, they can be applied simultaneously. We report the composition at batch size 1, so the values for our method differ from the batch size 4 results of Table~\ref{tab:main-result}.

\begin{table}[h]
\centering
\caption{Composition of our pruning method with R-KV KV-cache compression, at batch size 1 with a KV budget of 2048. Our method and the composition use 50\% target sparsity. Values are average accuracy over the five reasoning benchmarks.}
\label{tab:rkv}
\small
\begin{tabular}{l|ccc}
\toprule
\textbf{Model} & \textbf{Dense} & \textbf{Ours} & \textbf{R-KV + Ours} \\
\midrule
DeepSeek-R1-Distill-Qwen-7B  & 72.7 & 55.0 & 47.3 \\
DeepSeek-R1-Distill-Llama-8B & 71.3 & 37.6 & 31.4 \\
\bottomrule
\end{tabular}
\end{table}

As shown in Table~\ref{tab:rkv}, R-KV + Ours achieves accuracy close to that of our method alone while additionally providing R-KV's KV-cache memory savings. The accuracy drop in the composition arises from R-KV itself, which trades some accuracy for those savings. Practitioners can therefore stack the two methods to trade off KV-cache footprint against inference compute according to deployment constraints.

\section{Composition with Weight Quantization}
\label{sec:fp8_composition}

Pruning reduces the number of FFN neurons evaluated, whereas quantization reduces the cost of each parameter, so the two act on different axes and can be applied together. We verify that our method composes with FP8 weight quantization at 50\% target sparsity with batch size 4.

\begin{table}[h]
\centering
\caption{Composition of our pruning method with FP8 weight quantization at 50\% target sparsity with batch size 4. Values are average accuracy over the five reasoning benchmarks.}
\label{tab:fp8}
\small
\begin{tabular}{l|cc}
\toprule
\textbf{Model} & \textbf{Ours (BF16)} & \textbf{Ours (FP8)} \\
\midrule
DeepSeek-R1-Distill-Qwen-7B  & 54.0 & 54.3 \\
DeepSeek-R1-Distill-Llama-8B & 34.5 & 38.4 \\
\bottomrule
\end{tabular}
\end{table}

As shown in Table~\ref{tab:fp8}, FP8 weight quantization composes cleanly with our pruning at essentially no accuracy cost on DS-R1-Qwen-7B, and on DS-R1-Llama-8B accuracy even improves by 3.9 points. The two techniques are therefore not in competition: our method is an additional efficiency axis available on top of a quantized model, rather than an alternative to quantizing it.

\section{Full Performance Comparison}
\label{sec:appendix}

Table~\ref{tab:full-result} extends the main results (Table~\ref{tab:main-result}) to all batch sizes.
TEAL achieves strong performance at batch size 1 (68.3\% on DS-R1-Qwen-7B, 63.5\% on DS-R1-Llama-8B), but degrades sharply as batch size increases: on DS-R1-Qwen-7B, accuracy drops to 14.3\% at BS=4, 4.1\% at BS=8, and 6.8\% at BS=16; on DS-R1-Llama-8B, it falls to 10.4\% at BS=4, 2.4\% at BS=8, and 1.3\% at BS=16. This degradation is caused by the distribution shift when aggregating activations across multiple samples, which diverges from TEAL's single-sample calibration setting.
In contrast, our method maintains consistent performance across all batch sizes: on DS-R1-Qwen-7B, average accuracy ranges from 52.4\% to 55.0\% across BS=1--16, and on DS-R1-Llama-8B, from 30.4\% to 37.6\%.
Static pruning methods (Wanda and Griffin) are batch-size independent but exhibit poor performance on reasoning tasks overall, as they determine a fixed pruning pattern and cannot adapt to the evolving activation patterns during long chain-of-thought generation. Wanda achieves only 23.8\% and 5.3\%, while Griffin obtains 14.6\% and 10.4\% on DS-R1-Qwen-7B and DS-R1-Llama-8B respectively.

\begin{table}[!ht]
\centering
\caption{Pruning performance comparison at 50\% target sparsity across different batch sizes. Our method significantly outperforms all pruning baselines (Wanda, Griffin, TEAL). Dense, Wanda, and Griffin results are batch-size independent. Bold values indicate superior performance compared to TEAL at the same batch size.}
\label{tab:full-result}
\small
\resizebox{\columnwidth}{!}{%
\begin{tabular}{c|c|c|ccccc|c}
\toprule
\multirow{2}{*}{\textbf{Model}} & \multirow{2}{*}{\textbf{Method}} & \multirow{2}{*}{\textbf{Batch Size}} & \multicolumn{5}{c|}{\textbf{Tasks}} & \multirow{2}{*}{\textbf{AVG}} \\
\cmidrule(lr){4-8}
& & & \textbf{GSM8K} & \textbf{MATH500} & \textbf{MINERVA} & \textbf{AMC23} & \textbf{GPQA-DIAMOND} & \\
\midrule

\multirow{12}{*}{\textit{\makecell{DeepSeek-\\R1-Distill-\\Qwen-7B}}}
& Dense & -- & 92.0 & 91.8 & 39.7 & 87.5 & 52.5 & 72.7 \\
\cmidrule(l){2-9}
& Wanda & -- & 59.0 & 24.4 & 7.4 & 12.5 & 15.7 & 23.8 \\
& Griffin & -- & 22.0 & 14.8 & 11.0 & 10.0 & 15.2 & 14.6 \\
\cmidrule(l){2-9}
& \multirow{4}{*}{TEAL} & 1 & 91.0 & 86.8 & 40.1 & 80.0 & 43.4 & 68.3 \\
& & 4 & 30.0 & 11.2 & 4.8 & 10.0 & 15.7 & 14.3 \\
& & 8 & 9.0 & 3.2 & 3.3 & 0.0 & 5.0 & 4.1 \\
& & 16 & 0.0 & 1.6 & 0.7 & 30.0 & 1.5 & 6.8 \\
\cmidrule(l){2-9}
& \multirow{4}{*}{Ours} & 1 & 85.0 & 72.0 & 26.1 & 57.5 & 34.3 & 55.0 \\
& & 4 & 89.0 & 71.0 & 29.4 & 50.0 & 30.8 & 54.0 \\
& & 8 & 84.0 & 70.8 & 25.7 & 55.0 & 26.3 & 52.4 \\
& & 16 & 86.0 & 71.8 & 30.1 & 45.0 & 29.8 & 52.5 \\
\midrule

\multirow{12}{*}{\textit{\makecell{DeepSeek-\\R1-Distill-\\Llama-8B}}}
& Dense & -- & 96.0 & 91.0 & 33.8 & 90.0 & 45.5 & 71.3 \\
\cmidrule(l){2-9}
& Wanda & -- & 9.0 & 4.8 & 0.7 & 5.0 & 7.1 & 5.3 \\
& Griffin & -- & 16.0 & 11.8 & 4.8 & 2.5 & 16.7 & 10.4 \\
\cmidrule(l){2-9}
& \multirow{4}{*}{TEAL} & 1 & 95.0 & 81.8 & 30.9 & 70.0 & 39.9 & 63.5 \\
& & 4 & 28.0 & 5.4 & 2.2 & 0.0 & 16.2 & 10.4 \\
& & 8 & 2.0 & 3.8 & 2.6 & 2.5 & 1.0 & 2.4 \\
& & 16 & 1.0 & 3.2 & 2.2 & 0.0 & 0.0 & 1.3 \\
\cmidrule(l){2-9}
& \multirow{4}{*}{Ours} & 1 & 65.0 & 48.0 & 14.0 & 35.0 & 25.8 & 37.6 \\
& & 4 & 75.0 & 38.8 & 12.1 & 25.0 & 21.7 & 34.5 \\
& & 8 & 69.0 & 39.0 & 8.8 & 12.5 & 22.7 & 30.4 \\
& & 16 & 65.0 & 39.6 & 9.6 & 25.0 & 16.7 & 31.2 \\
\bottomrule
\end{tabular}%
}
\end{table}

\section{Generation Throughput at 50\% Target Sparsity}
\label{sec:target_sparsity_speed}

All throughput experiments in this section are conducted on NVIDIA H200 GPUs.
Tables~\ref{tab:speed-comparison-qwen} and \ref{tab:speed-comparison-llama} report the throughput at 50\% target sparsity. Note that the main text reports throughput under actual sparsity on an NVIDIA H100 GPU (Figure~\ref{fig:speed-actual-sparsity}); this section provides the complementary comparison at identical target sparsity on an NVIDIA H200 GPU.

Griffin achieves the highest speedup across all settings (e.g., $1.61\times$ at BS=1 on DS-R1-Qwen-7B) because it applies a fixed pruning mask without adaptive overhead; however, as shown in Table~\ref{tab:full-result}, its accuracy on reasoning tasks is severely degraded.
At BS=1, TEAL achieves higher speedup than our method ($1.45\times$ vs.\ $1.32\times$ on DS-R1-Qwen-7B), as its unstructured sparsity via near-zero activation skipping is effective in the memory-bound regime. However, as batch size increases, all methods' speedup ratios converge: at BS=16, TEAL, Griffin, and our method achieve $1.10\times$, $1.27\times$, and $1.10\times$ respectively on DS-R1-Qwen-7B.

\begin{table}[!ht]
\centering
\caption{Generation throughput (tokens/sec) and speedup comparison on DeepSeek-R1-Distill-Qwen-7B at 50\% target sparsity relative to Dense baseline across different batch sizes on NVIDIA H200 GPU.}
\label{tab:speed-comparison-qwen}
\small
\begin{tabular}{l|cccc}
\toprule
\textbf{Method} & \textbf{BS=1} & \textbf{BS=4} & \textbf{BS=8} & \textbf{BS=16} \\
\midrule
Dense & 200.9 (1.00$\times$) & 813.0 (1.00$\times$) & 1596.0 (1.00$\times$) & 3040.5 (1.00$\times$) \\
TEAL & 290.7 (1.45$\times$) & 971.1 (1.19$\times$) & 1769.1 (1.11$\times$) & 3335.9 (1.10$\times$) \\
GRIFFIN & 323.5 (1.61$\times$) & 1084.5 (1.33$\times$) & 2111.0 (1.32$\times$) & 3854.3 (1.27$\times$) \\
Ours & 265.6 (1.32$\times$) & 925.3 (1.14$\times$) & 1794.7 (1.12$\times$) & 3331.8 (1.10$\times$) \\
\bottomrule
\end{tabular}
\end{table}

\FloatBarrier

On DS-R1-Llama-8B, our method maintains a consistent $1.15\times$ speedup across BS=4, 8, and 16, slightly outperforming TEAL ($1.10\text{--}1.11\times$) at BS=8 and BS=16.

\begin{table}[!ht]
\centering
\caption{Generation throughput (tokens/sec) and speedup comparison on DeepSeek-R1-Distill-Llama-8B at 50\% target sparsity relative to Dense baseline across different batch sizes on NVIDIA H200 GPU.}
\label{tab:speed-comparison-llama}
\small
\begin{tabular}{l|cccc}
\toprule
\textbf{Method} & \textbf{BS=1} & \textbf{BS=4} & \textbf{BS=8} & \textbf{BS=16} \\
\midrule
Dense & 175.0 (1.00$\times$) & 755.2 (1.00$\times$) & 1471.7 (1.00$\times$) & 2713.0 (1.00$\times$) \\
TEAL & 263.8 (1.51$\times$) & 837.2 (1.11$\times$) & 1618.2 (1.10$\times$) & 2997.4 (1.10$\times$) \\
GRIFFIN & 269.5 (1.54$\times$) & 1010.5 (1.34$\times$) & 1957.1 (1.33$\times$) & 3747.6 (1.38$\times$) \\
Ours & 227.3 (1.30$\times$) & 868.1 (1.15$\times$) & 1685.9 (1.15$\times$) & 3124.9 (1.15$\times$) \\
\bottomrule
\end{tabular}
\end{table}

\section{Generation Throughput across Batch Sizes}
\label{sec:actual_sparsity_speed_all_bs}

\begin{figure}[!ht]
    \centering
    \includegraphics[width=0.95\linewidth]{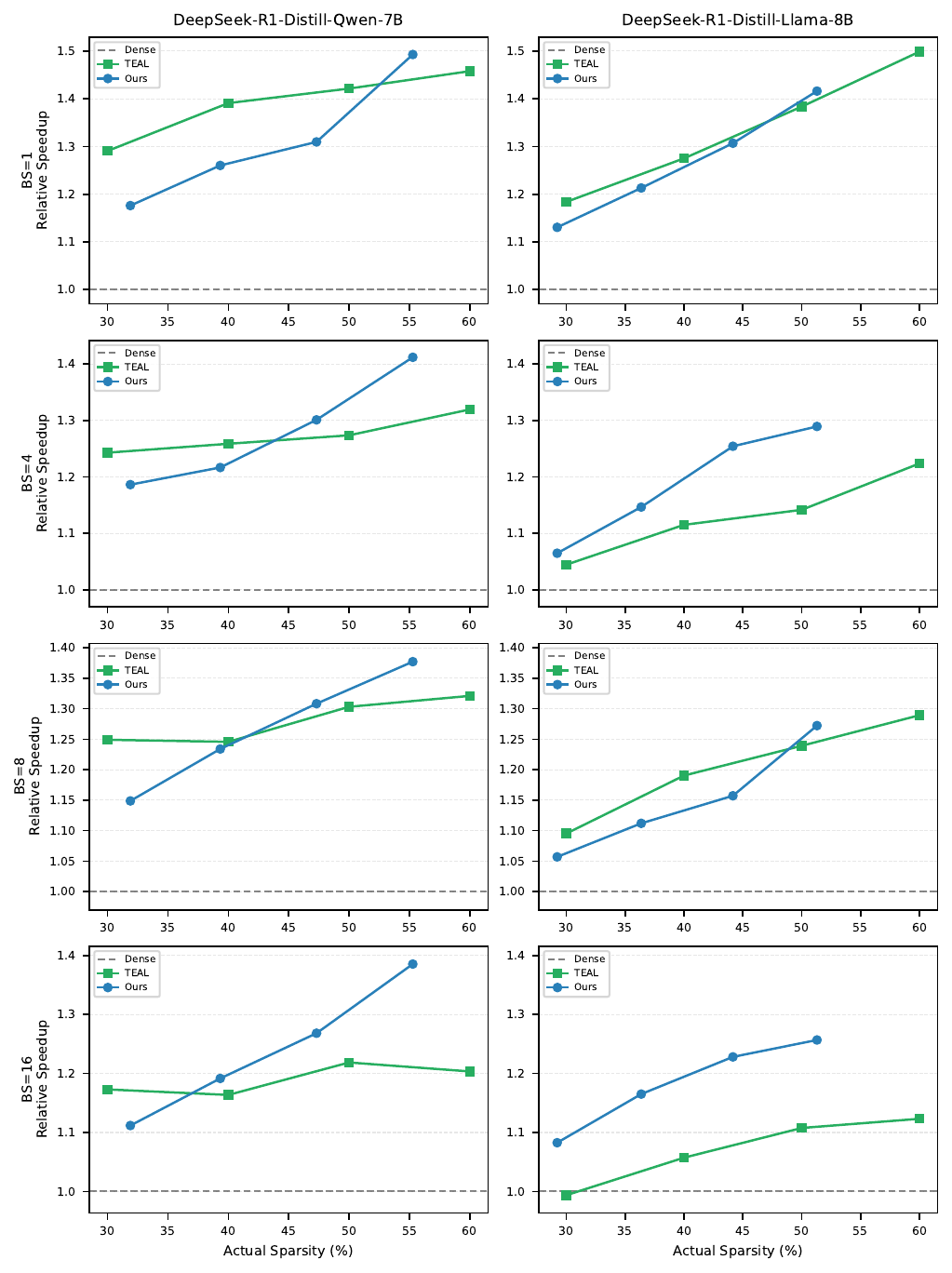}
    \caption{Relative generation throughput speedup vs.\ actual sparsity across batch sizes on NVIDIA H100 GPU. Each row corresponds to a different batch size. (Left) DeepSeek-R1-Distill-Qwen-7B, (Right) DeepSeek-R1-Distill-Llama-8B.}
    \label{fig:speed-actual-sparsity-all-bs}
\end{figure}

Figure~\ref{fig:speed-actual-sparsity-all-bs} shows the relative throughput speedup versus actual sparsity across batch sizes 1, 4, 8, and 16 on an NVIDIA H100 GPU. At batch size 1, where inference is memory-bound, TEAL's sparse kernel achieves higher speedup than our method across all sparsity levels. However, as batch size increases, our method increasingly outperforms TEAL. At batch size 8, our method surpasses TEAL at higher sparsity levels on both models. At batch size 16, our method achieves higher speedup than TEAL across nearly all sparsity levels. This trend confirms that structurally reducing parameters via smaller weight matrices becomes more effective than skipping near-zero activations as the workload shifts from memory-bound to compute-bound.

\section{Speed-Accuracy Trade-off}

\begin{figure}[!ht]
\centering
\includegraphics[width=\linewidth]{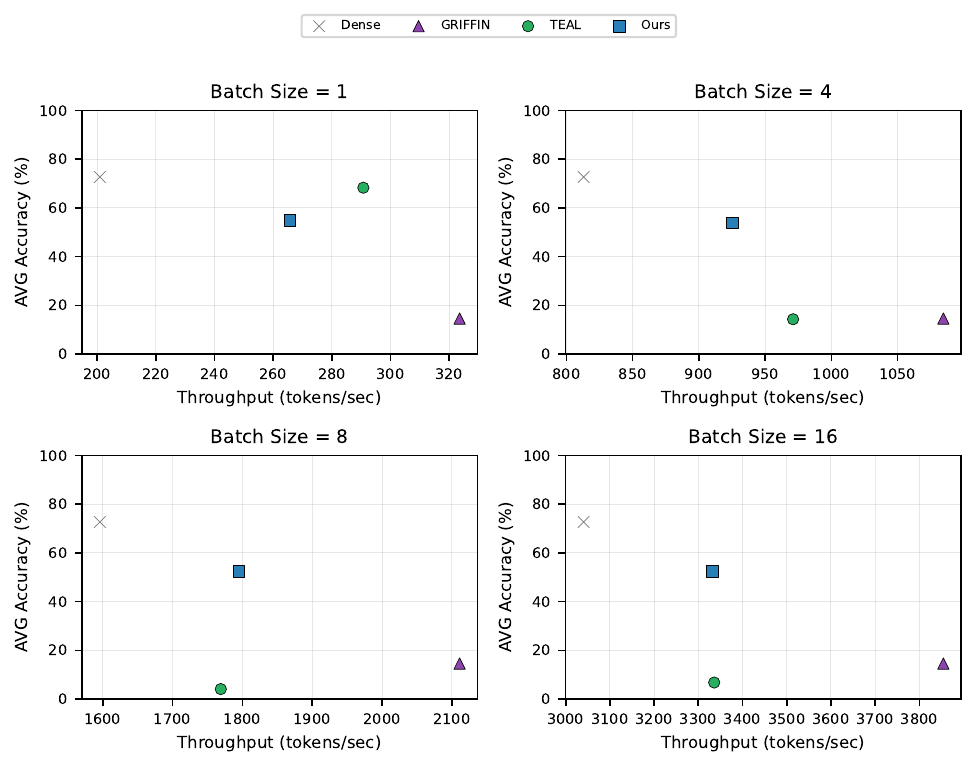}
\caption{Throughput versus average accuracy trade-off on DeepSeek-R1-Distill-Qwen-7B across different batch sizes at 50\% target sparsity on NVIDIA H200 GPU. Each subplot shows the speed-accuracy relationship for a specific batch size. Our method maintains consistent accuracy across batch sizes while achieving competitive throughput, whereas TEAL's accuracy degrades significantly as batch size increases.}
\label{fig:speed-vs-accuracy-qwen}
\end{figure}

\begin{figure}[!ht]
\centering
\includegraphics[width=\linewidth]{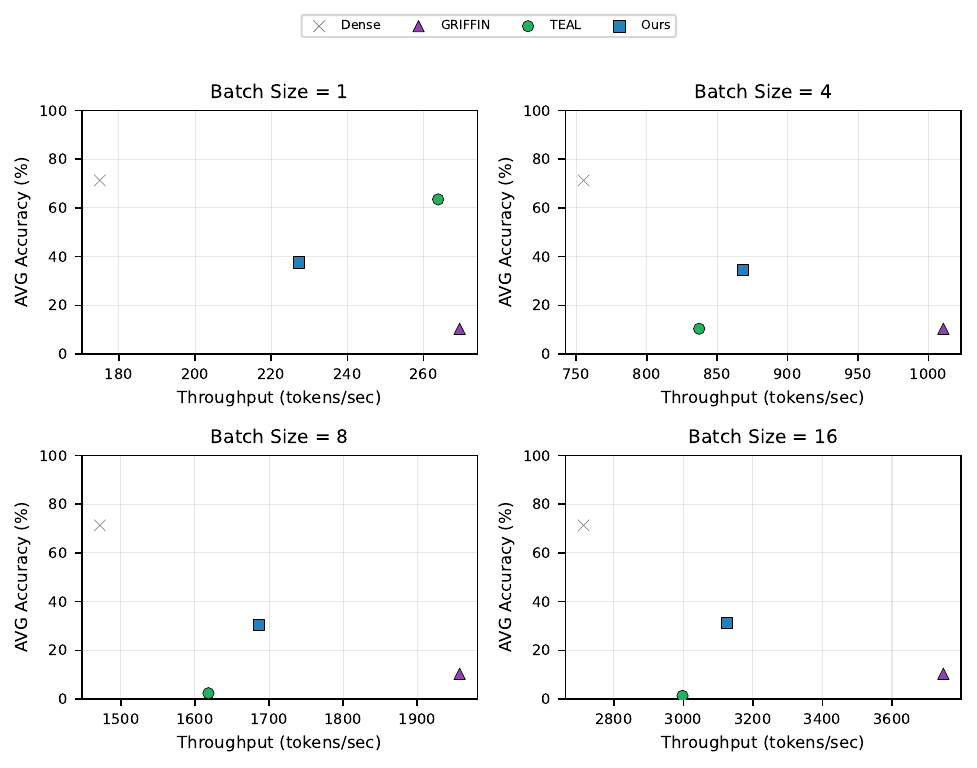}
\caption{Throughput versus average accuracy trade-off on DeepSeek-R1-Distill-Llama-8B across different batch sizes at 50\% target sparsity on NVIDIA H200 GPU. Each subplot shows the speed-accuracy relationship for a specific batch size. Our method maintains consistent accuracy across batch sizes while achieving competitive throughput, whereas TEAL's accuracy degrades significantly as batch size increases.}
\label{fig:speed-vs-accuracy-llama}
\end{figure}

Figures~\ref{fig:speed-vs-accuracy-qwen} and \ref{fig:speed-vs-accuracy-llama} visualize the throughput versus accuracy trade-off across batch sizes. At BS=1, TEAL occupies a favorable position with high accuracy (68.3\% on DS-R1-Qwen-7B, 63.5\% on DS-R1-Llama-8B) and high throughput. However, as batch size increases, TEAL loses both accuracy and relative speedup, dropping to 6.8\% and 1.3\% average accuracy at BS=16 on DS-R1-Qwen-7B and DS-R1-Llama-8B respectively. In contrast, our method maintains consistent accuracy across all batch sizes (55.0\% to 52.5\% on DS-R1-Qwen-7B, 37.6\% to 30.4\% on DS-R1-Llama-8B) with competitive throughput, demonstrating a more favorable trade-off in batched deployment scenarios.

\section{Dataset and Model Details}

We summarize the evaluation datasets and models used in this work below.

\subsection{Dataset Details}
We report the split, number of examples, and license for each dataset.

\begin{itemize}
  \item \textbf{TinyGSM8K}:
  split = test, \#examples = 100, license = MIT.

  \item \textbf{MATH500}:
  split = test, \#examples = 500, license = MIT.

  \item \textbf{MINERVA Math}:
  split = test, \#examples = 272, license = MIT.

  \item \textbf{AMC23}:
  split = test, \#examples = 40, license = unspecified (MAA copyrighted).

  \item \textbf{GPQA-DIAMOND}:
  split = test, \#examples = 198, license = CC BY 4.0.
\end{itemize}

\subsection{Model Details}
We evaluate on two reasoning models and two additional models, and use a consistent zero-shot inference protocol.

\begin{itemize}
  \item \textbf{DeepSeek-R1-Distill-Qwen-7B}:
  a distilled reasoning model (7B). License = MIT (model weights/repository).
  Base lineage = Qwen-2.5 series (Apache 2.0).
  Evaluation uses the official DeepSeek-R1 chat template, greedy decoding with temperature $0$ (zero-shot),
  accuracy as the metric, and maximum generation length of 16,000 tokens.

  \item \textbf{DeepSeek-R1-Distill-Llama-8B}:
  a distilled reasoning model (8B). License = MIT (model weights/repository).
  Base lineage = Llama-3.1-8B-Base (Llama 3.1 license).
  We use the same evaluation protocol: official DeepSeek-R1 chat template, greedy decoding with temperature $0$ (zero-shot),
  accuracy as the metric, and maximum generation length of 16,000 tokens.

  \item \textbf{Qwen3-1.7B}:
  a reasoning model (1.7B) from the Qwen3 series~\citep{qwen3}. License = Apache 2.0.
  Evaluation uses the Qwen3 chat template with \texttt{enable\_thinking=True}, greedy decoding with temperature $0$ (zero-shot),
  accuracy as the metric, and maximum generation length of 16,000 tokens.

  \item \textbf{Qwen3-8B}:
  a reasoning model (8B) from the Qwen3 series~\citep{qwen3}. License = Apache 2.0.
  We use the same evaluation protocol as Qwen3-1.7B: Qwen3 chat template with \texttt{enable\_thinking=True}, greedy decoding with temperature $0$ (zero-shot),
  accuracy as the metric, and maximum generation length of 16,000 tokens.
\end{itemize}


\section{Computational Overhead}
\label{sec:flops_overhead}

Table~\ref{tab:flops} reports the per-token FLOPs for our adaptive mask computation. All measurements are conducted in Float16. The additional FLOPs introduced by adaptive mask computation account for less than 0.002\% of the dense MLP computation, while our method reduces approximately 45\% of the MLP computation, which is close to the target sparsity.

\begin{table}[!ht]
\centering
\caption{Per-token FLOPs analysis. Overhead refers to adaptive mask computation.}
\label{tab:flops}
\small
\begin{tabular}{l|cccc}
\toprule
\textbf{Model} & \textbf{Dense MLP} & \textbf{Ours MLP} & \textbf{Overhead} & \textbf{Reduction} \\
\midrule
DeepSeek-R1-Distill-Qwen-7B  & 11.41B & 6.27B & 212.18K & 45.0\% \\
DeepSeek-R1-Distill-Llama-8B & 11.27B & 6.20B & 183.51K & 45.0\% \\
\bottomrule
\end{tabular}
\end{table}

\section{Non-Reasoning Task Performance}
\label{sec:nonreasoning}

Table~\ref{tab:nonreasoning} reports performance on non-reasoning benchmarks at 50\% target sparsity. Our method preserves general task performance within approximately 1\% of the dense baseline across all models: the average accuracy gap is 0.5\% on DS-R1-Qwen-7B (63.6\% vs.\ 64.1\%), 0.7\% on DS-R1-Llama-8B (71.9\% vs.\ 72.6\%), 1.2\% on Qwen3-1.7B (65.2\% vs.\ 66.4\%), and 1.0\% on Qwen3-8B (76.0\% vs.\ 77.0\%). Our method also outperforms TEAL on average across all four models, with both methods showing negligible degradation compared to the dense baseline on these shorter-output tasks.

\begin{table}[!ht]
\centering
\caption{Non-reasoning benchmark performance at 50\% target sparsity. Bold indicates best among pruning methods.}
\label{tab:nonreasoning}
\small
\resizebox{\columnwidth}{!}{%
\begin{tabular}{ll|cccccc|c}
\toprule
\textbf{Model} & \textbf{Method} & \textbf{HellaSwag} & \textbf{PIQA} & \textbf{COPA} & \textbf{ARC-E} & \textbf{ARC-C} & \textbf{BoolQ} & \textbf{Avg} \\
\midrule
\multirow{3}{*}{\textit{\makecell[l]{DeepSeek-R1-\\Distill-Qwen-7B}}}
& Dense & 59.7 & 70.2 & 70.0 & 67.1 & 40.3 & 77.6 & 64.1 \\
\cmidrule(lr){2-9}
& TEAL  & \textbf{60.0} & 69.6 & \textbf{71.0} & \textbf{65.3} & 38.7 & 74.9 & 63.3 \\
& Ours  & 59.6 & \textbf{70.2} & \textbf{71.0} & 65.2 & \textbf{39.6} & \textbf{76.1} & \textbf{63.6} \\
\midrule
\multirow{3}{*}{\textit{\makecell[l]{DeepSeek-R1-\\Distill-Llama-8B}}}
& Dense & 74.8 & 77.0 & 90.0 & 69.9 & 40.7 & 83.5 & 72.6 \\
\cmidrule(lr){2-9}
& TEAL  & \textbf{74.7} & \textbf{77.1} & 87.0 & \textbf{69.3} & 39.9 & 82.1 & 71.7 \\
& Ours  & \textbf{74.7} & 76.7 & \textbf{88.0} & 68.8 & \textbf{40.9} & \textbf{82.3} & \textbf{71.9} \\
\midrule
\multirow{3}{*}{\textit{Qwen3-1.7B}}
& Dense & 60.5 & 72.4 & 75.0 & 72.9 & 40.4 & 77.4 & 66.4 \\
\cmidrule(lr){2-9}
& TEAL  & \textbf{60.4} & 72.2 & \textbf{75.0} & \textbf{71.4} & 37.9 & 73.4 & 65.0 \\
& Ours  & 60.3 & \textbf{72.8} & \textbf{75.0} & 70.5 & \textbf{38.4} & \textbf{73.9} & \textbf{65.2} \\
\midrule
\multirow{3}{*}{\textit{Qwen3-8B}}
& Dense & 74.9 & 76.6 & 85.0 & 83.3 & 55.5 & 86.6 & 77.0 \\
\cmidrule(lr){2-9}
& TEAL  & 74.7 & 76.4 & 84.0 & \textbf{82.6} & \textbf{54.7} & 80.2 & 75.4 \\
& Ours  & \textbf{74.9} & \textbf{76.8} & \textbf{85.0} & 81.0 & 52.9 & \textbf{85.5} & \textbf{76.0} \\
\bottomrule
\end{tabular}%
}
\end{table}

\FloatBarrier
\section{Extended Periodicity Analysis}
\label{sec:qwen3_periodicity}

Table~\ref{tab:periodicity_qwen3} extends the periodicity analysis (Table~\ref{tab:periodicity} in the main text) to all evaluated models. Based on the observed median activation periods, we set $T_{\text{trans}}=20$ for DS-R1-Qwen-7B, DS-R1-Llama-8B, and Qwen3-8B, and $T_{\text{trans}}=10$ for Qwen3-1.7B (with $T_{\text{init}}=64$ and $T_E=2$).

\begin{table}[!ht]
\centering
\caption{Periodicity analysis of important neuron activations across all evaluated models (mean $\pm$ std across 10 samples $\times$ 5 benchmarks).}
\label{tab:periodicity_qwen3}
\small
\begin{tabular}{l|cc}
\toprule
\textbf{Model} & \textbf{Median Period} & \textbf{Re-firing (\%)} \\
\midrule
DeepSeek-R1-Distill-Qwen-7B  & 22.8 $\pm$ 1.9 & 76.3 $\pm$ 2.8 \\
DeepSeek-R1-Distill-Llama-8B & 20.8 $\pm$ 2.9 & 71.9 $\pm$ 2.2 \\
\midrule
Qwen3-1.7B & 11.8 $\pm$ 0.8 & 80.7 $\pm$ 3.1 \\
Qwen3-8B   & 17.8 $\pm$ 1.3 & 76.8 $\pm$ 2.9 \\
\bottomrule
\end{tabular}
\end{table}

\section{Extended Ablation}
\label{sec:design_ablation_full}

Table~\ref{tab:design_ablation_full} provides the full per-benchmark breakdown for the design choice ablation summarized in Table~\ref{tab:design_ablation}. The upper section compares batch aggregation methods (max vs.\ mean), and the lower section evaluates the effect of activation memory (on vs.\ off).

For aggregation, max and mean show task-dependent strengths: in cross-model average, max outperforms mean on GSM8K (82.0 vs.\ 73.5) and AMC23 (37.5 vs.\ 35.0), while mean is stronger on MATH500 (58.1 vs.\ 54.9) and GPQA (27.8 vs.\ 26.3). The cross-model averages differ by 1.2 points (44.3 vs.\ 43.1), which we do not consider sufficient to claim that either operator is superior; the per-model winner is likewise split (Section~\ref{sec:ablation}).
For activation memory, enabling memory consistently improves average accuracy by 8.7 points on DS-R1-Qwen-7B (54.0 vs.\ 45.3) and 9.0 points on DS-R1-Llama-8B (34.5 vs.\ 25.5). The gains are particularly large on GSM8K and MATH500. We note that on GPQA-DIAMOND, the no-memory variant slightly outperforms the memory variant on DS-R1-Llama-8B (29.3 vs.\ 21.7), suggesting that accumulated activation memory may occasionally over-retain neurons from earlier phases on certain tasks.

\begin{table}[!ht]
\centering
\caption{Per-benchmark performance comparison of aggregation methods (mean/max) and activation memory (on/off) at 50\% target sparsity with batch size 4. Bold indicates best per model.}
\label{tab:design_ablation_full}
\small
\resizebox{\columnwidth}{!}{%
\begin{tabular}{l|ccccc|c}
\toprule
\textbf{Setting} & \textbf{GSM8K} & \textbf{MATH500} & \textbf{MINERVA} & \textbf{AMC23} & \textbf{GPQA-DIAMOND} & \textbf{AVG} \\
\midrule
\multicolumn{7}{l}{\textit{DeepSeek-R1-Distill-Qwen-7B}} \\
\midrule
Max (Ours) & \textbf{89.0} & 71.0 & \textbf{29.4} & 50.0 & \textbf{30.8} & 54.0 \\
Mean       & 83.0 & \textbf{74.0} & \textbf{29.4} & \textbf{57.5} & 30.3 & \textbf{54.8} \\
\cmidrule(lr){1-7}
Memory (Ours)    & \textbf{89.0} & \textbf{71.0} & \textbf{29.4} & \textbf{50.0} & \textbf{30.8} & \textbf{54.0} \\
No Memory        & 76.0 & 63.6 & 21.7 & 37.5 & 27.8 & 45.3 \\
\midrule
\multicolumn{7}{l}{\textit{DeepSeek-R1-Distill-Llama-8B}} \\
\midrule
Max (Ours) & \textbf{75.0} & 38.8 & 12.1 & \textbf{25.0} & 21.7 & \textbf{34.5} \\
Mean       & 64.0 & \textbf{42.2} & \textbf{13.2} & 12.5 & \textbf{25.3} & 31.4 \\
\cmidrule(lr){1-7}
Memory (Ours)    & \textbf{75.0} & \textbf{38.8} & \textbf{12.1} & \textbf{25.0} & 21.7 & \textbf{34.5} \\
No Memory        & 46.0 & 26.2 & 8.5 & 17.5 & \textbf{29.3} & 25.5 \\
\midrule
\multicolumn{7}{l}{\textit{Cross-model Average}} \\
\midrule
Max (Ours) & \textbf{82.0} & 54.9 & 20.8 & \textbf{37.5} & 26.3 & \textbf{44.3} \\
Mean       & 73.5 & \textbf{58.1} & \textbf{21.3} & 35.0 & \textbf{27.8} & 43.1 \\
\cmidrule(lr){1-7}
Memory (Ours)    & \textbf{82.0} & \textbf{54.9} & \textbf{20.8} & \textbf{37.5} & 26.3 & \textbf{44.3} \\
No Memory        & 61.0 & 44.9 & 15.1 & 27.5 & \textbf{28.6} & 35.4 \\
\bottomrule
\end{tabular}%
}
\end{table}

\section{Hyperparameter Ablation}
\label{sec:hyperparameter_ablation}

Table~\ref{tab:parameter_ablation} presents hyperparameter ablation results on MATH500 with batch size 4. Increasing $T_{\text{init}}$ (initial dense steps) from 0 to 128 improves accuracy by 2.8 points with minimal speed impact, as it captures post-prompt activation shifts. Increasing $T_E$ (exploration steps) from 1 to 4 yields a 9.8 point accuracy gain but reduces throughput from 941.0 to 896.8 tokens/sec. Conversely, extending $T_{\text{trans}}$ (update period) from 10 to 30 degrades accuracy by 7.8 points while increasing throughput from 806.3 to 958.7 tokens/sec. In summary, $T_{\text{init}}$ primarily affects accuracy without speed penalty, while $T_E$ and $T_{\text{trans}}$ present accuracy-speed trade-offs. Hyperparameter values can be selected based on deployment requirements.

We additionally evaluate the most frequent update setting ($T_{\text{init}}\!=\!0, T_E\!=\!1, T_{\text{trans}}\!=\!2$), which alternates one exploration step with one pruning step. It achieves 88.2 accuracy on DS-R1-Qwen-7B and 77.6 on DS-R1-Llama-8B for MATH500. This configuration, however, recomputes the top-$k$ mask every other step and therefore pays the selection cost far more often than a longer update period does; we do not report a throughput figure for it, and we do not adopt it, since the speedup that motivates pruning is what such frequent updates forfeit.

\begin{table}[!ht]
    \centering
    \caption{Hyperparameter ablation study on DeepSeek-R1-Distill-Qwen-7B using the MATH500 dataset with batch size 4 on NVIDIA H200 GPU, examining the effects of $T_{\text{init}}$ (initial dense steps), $T_E$ (exploration steps), and $T_{\text{trans}}$ (update period) on accuracy and inference speed.}
    \label{tab:parameter_ablation}
    \small
    \begin{tabular}{ccc|cc}
    \toprule
    \textbf{$T_{\text{init}}$} & \textbf{$T_E$} & \textbf{$T_{\text{trans}}$} & \textbf{Accuracy} & \textbf{Speed (token/sec)} \\
    \midrule
    0   & 2 & 20 & 68.6 & 925.3 \\
    32  & 2 & 20 & 70.8 & 925.3 \\
    64  & 2 & 20 & 71.0 & 925.3 \\
    128 & 2 & 20 & 71.4 & 925.3 \\
    \midrule
    64 & 1 & 20 & 67.4 & 941.0 \\
    64 & 2 & 20 & 71.0 & 925.3 \\
    64 & 4 & 20 & 77.2 & 896.8 \\
    \midrule
    64 & 2 & 10 & 77.4 & 806.3 \\
    64 & 2 & 20 & 71.0 & 925.3 \\
    64 & 2 & 30 & 69.6 & 958.7 \\
    \midrule
    0  & 1 & 2  & 88.2 & - \\
    \bottomrule
    \end{tabular}
\end{table}

\section{Threshold-Based Pruning}
\label{sec:threshold_ablation}

Table~\ref{tab:threshold_ablation} evaluates threshold-based pruning, where we calibrate a fixed activation threshold using 20,248 tokens from the C4 dataset, following the same protocol as TEAL~\citep{teal}. At 50\% target sparsity, this approach results in severe over-pruning with near-zero accuracy across all benchmarks on both models, confirming that a static threshold is inadequate for our adaptive pruning framework.

\begin{table}[!ht]
\centering
\caption{Threshold-based pruning at 50\% target sparsity with batch size 4. The activation threshold is calibrated on C4.}
\label{tab:threshold_ablation}
\small
\resizebox{\columnwidth}{!}{%
\begin{tabular}{l|l|ccccc|c}
\toprule
\textbf{Model} & \textbf{Method} & \textbf{GSM8K} & \textbf{MATH500} & \textbf{MINERVA} & \textbf{AMC23} & \textbf{GPQA-DIAMOND} & \textbf{AVG} \\
\midrule
\multirow{3}{*}{\textit{\makecell[l]{DeepSeek-R1-\\Distill-Qwen-7B}}}
& Dense & 92.0 & 91.8 & 39.7 & 87.5 & 52.5 & 72.7 \\
& Ours (top-$k$) & 89.0 & 71.0 & 29.4 & 50.0 & 30.8 & 54.0 \\
& Ours (threshold) & 3.0 & 3.2 & 2.9 & 0.0 & 1.5 & 2.1 \\
\midrule
\multirow{3}{*}{\textit{\makecell[l]{DeepSeek-R1-\\Distill-Llama-8B}}}
& Dense & 96.0 & 91.0 & 33.8 & 90.0 & 45.5 & 71.3 \\
& Ours (top-$k$) & 75.0 & 38.8 & 12.1 & 25.0 & 21.7 & 34.5 \\
& Ours (threshold) & 3.0 & 2.8 & 0.7 & 2.5 & 0.0 & 1.8 \\
\bottomrule
\end{tabular}%
}
\end{table}

\section{Extended Pruning Performance}
\label{sec:qwen3_results_pruning}

Table~\ref{tab:qwen3_pruning} reports pruning performance on additional models at 50\% target sparsity with batch size 4. Under the hyperparameter settings guided by the periodicity analysis (Appendix~\ref{sec:qwen3_periodicity}), our method achieves substantially better performance than TEAL across multiple reasoning tasks. On Qwen3-1.7B, our method achieves 44.0\% average accuracy compared to TEAL's 2.2\%, an improvement of 41.8 points. On Qwen3-8B, the gap is even larger: our method attains 52.7\% versus TEAL's 7.2\%, a 45.5 point improvement. Notably, TEAL's performance collapses nearly completely on both Qwen3 models in batched settings (e.g., 0.0\% on AMC23 and GPQA for Qwen3-1.7B), while our method retains meaningful accuracy across all benchmarks. For DS-R1-Qwen-7B and DS-R1-Llama-8B, batch size 4 results are also included in the full comparison (Table~\ref{tab:full-result}).

\begin{table}[!ht]
\centering
\caption{Pruning performance comparison at 50\% target sparsity with batch size 4. Bold indicates best among pruning methods.}
\label{tab:qwen3_pruning}
\small
\resizebox{\columnwidth}{!}{%
\begin{tabular}{l|l|ccccc|c}
\toprule
\textbf{Model} & \textbf{Method} & \textbf{GSM8K} & \textbf{MATH500} & \textbf{MINERVA} & \textbf{AMC23} & \textbf{GPQA-DIAMOND} & \textbf{AVG} \\
\midrule
\multirow{3}{*}{\textit{\makecell[l]{DeepSeek-R1-\\Distill-Qwen-7B}}}
& Dense & 92.0 & 91.8 & 39.7 & 87.5 & 52.5 & 72.7 \\
\cmidrule(lr){2-8}
& TEAL  & 30.0 & 11.2 & 4.8 & 10.0 & 15.7 & 14.3 \\
& Ours  & \textbf{89.0} & \textbf{71.0} & \textbf{29.4} & \textbf{50.0} & \textbf{30.8} & \textbf{54.0} \\
\midrule
\multirow{3}{*}{\textit{\makecell[l]{DeepSeek-R1-\\Distill-Llama-8B}}}
& Dense & 96.0 & 91.0 & 33.8 & 90.0 & 45.5 & 71.3 \\
\cmidrule(lr){2-8}
& TEAL  & 28.0 & 5.4 & 2.2 & 0.0 & 16.2 & 10.4 \\
& Ours  & \textbf{75.0} & \textbf{38.8} & \textbf{12.1} & \textbf{25.0} & \textbf{21.7} & \textbf{34.5} \\
\midrule
\multirow{3}{*}{\textit{Qwen3-1.7B}}
& Dense & \textbf{95.0} & \textbf{85.6} & \textbf{32.0} & \textbf{75.0} & \textbf{35.4} & \textbf{64.6} \\
\cmidrule(lr){2-8}
& TEAL  & 7.0 & 2.8 & 1.1 & 0.0 & 0.0 & 2.2 \\
& Ours  & \textbf{82.0} & \textbf{61.0} & \textbf{18.0} & \textbf{32.5} & \textbf{26.3} & \textbf{44.0} \\
\midrule
\multirow{3}{*}{\textit{Qwen3-8B}}
& Dense & \textbf{95.0} & \textbf{91.4} & \textbf{41.5} & \textbf{92.5} & \textbf{54.0} & \textbf{74.9} \\
\cmidrule(lr){2-8}
& TEAL  & 24.0 & 4.4 & 4.0 & 2.5 & 1.0 & 7.2 \\
& Ours  & \textbf{93.0} & \textbf{61.6} & \textbf{25.4} & \textbf{40.0} & \textbf{43.4} & \textbf{52.7} \\
\bottomrule
\end{tabular}%
}
\end{table}

\section{Extensive Activation Visualization}
\label{sec:activation_visualization}

\begin{figure}[!ht]
\centering
\includegraphics[width=\textwidth]{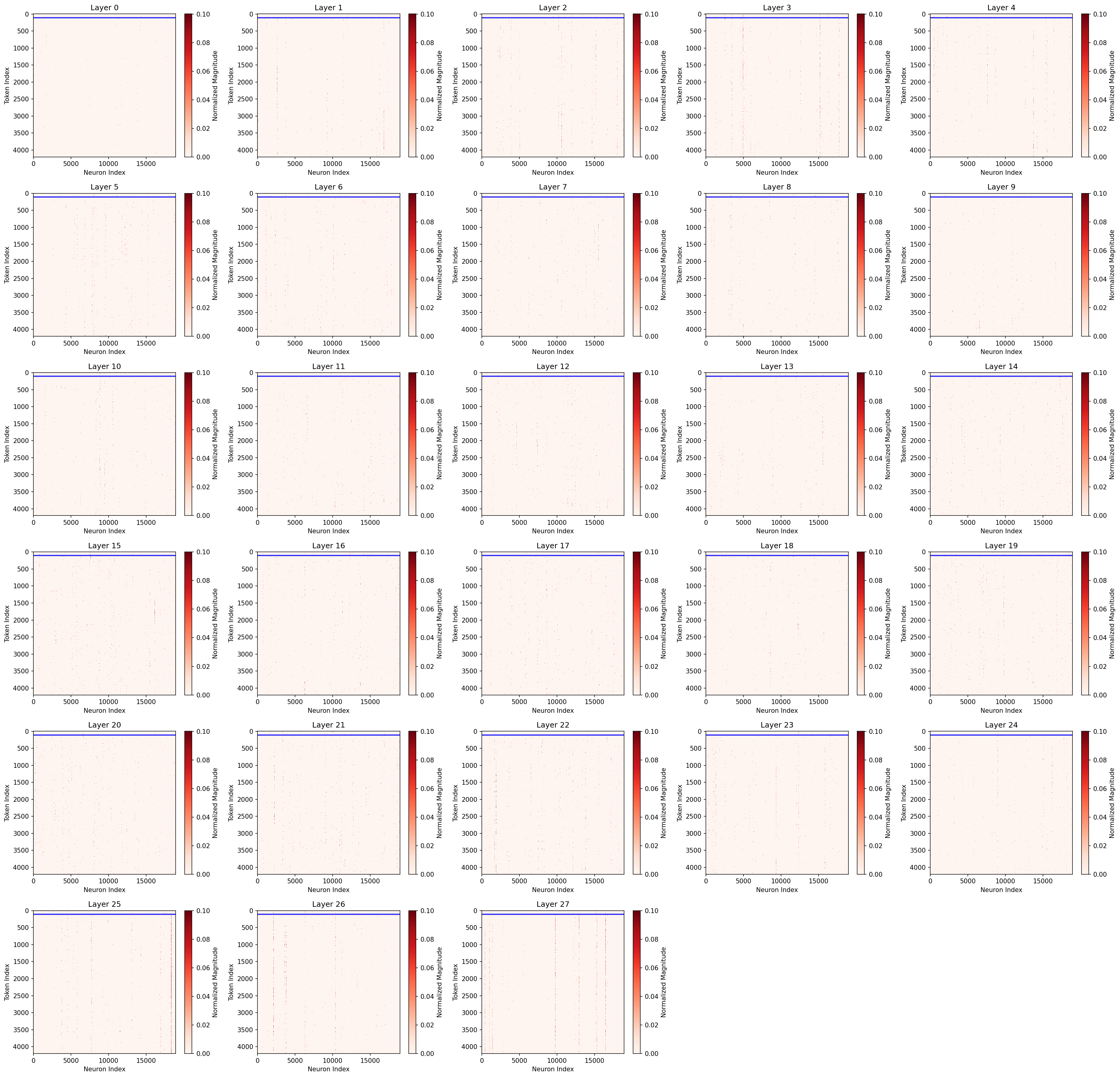}
\caption{Visualization of feedforward activations $\mathbf{Z}$ across all layers ($\bar{\mathbf{Z}} > 0.05$) of DeepSeek-R1-Distill-Qwen-7B on MATH500. The x-axis represents neurons and the y-axis represents token indices from top to bottom. The blue horizontal line indicates the last position of the prompt.}
\label{fig:activation-viz-qwen-math500}
\end{figure}

\begin{figure}[!ht]
\centering
\includegraphics[width=\textwidth]{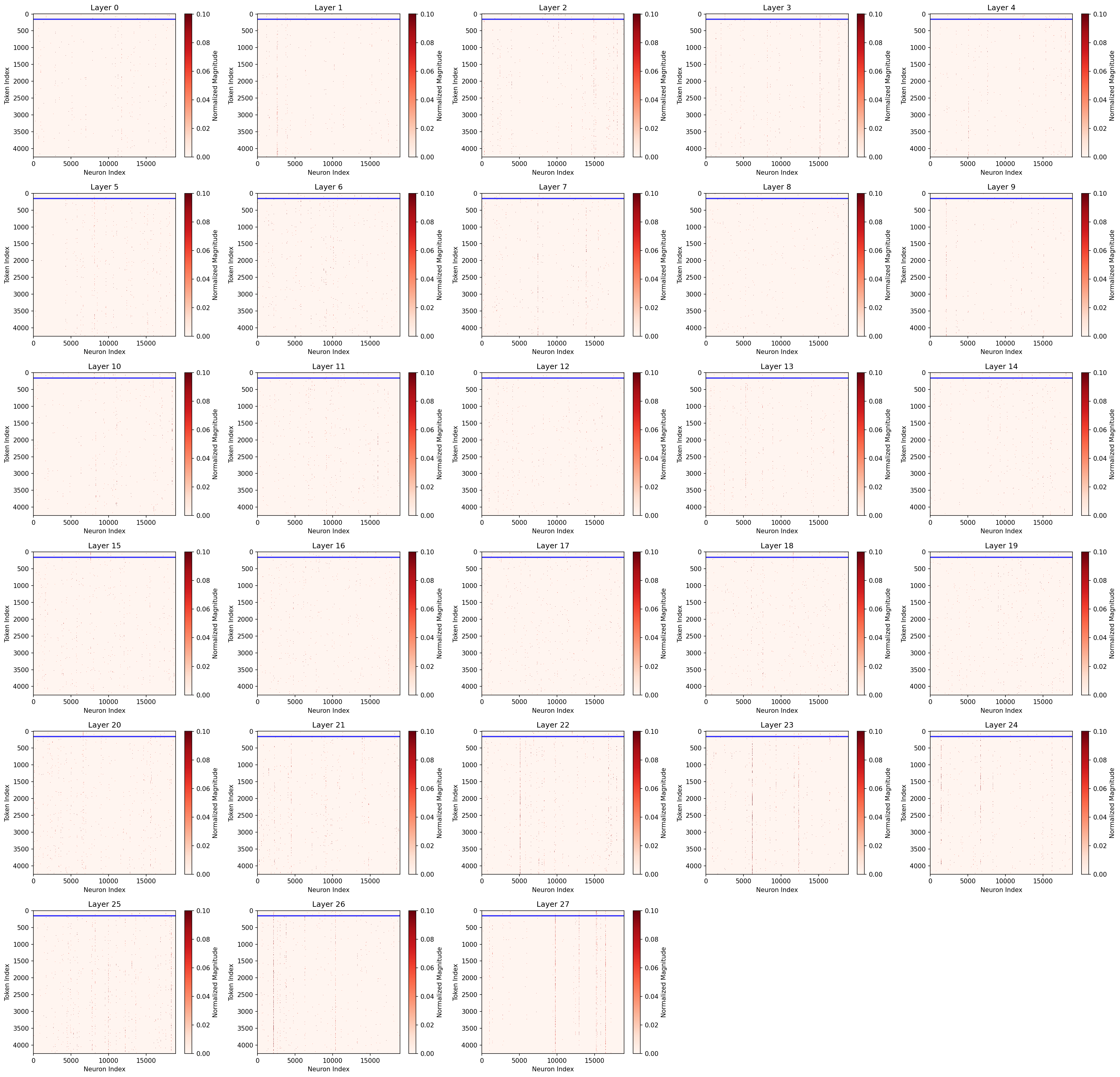}
\caption{Visualization of feedforward activations $\mathbf{Z}$ across all layers ($\bar{\mathbf{Z}} > 0.05$) of DeepSeek-R1-Distill-Qwen-7B on GPQA-DIAMOND. The x-axis represents neurons and the y-axis represents token indices from top to bottom. The blue horizontal line indicates the last position of the prompt.}
\label{fig:activation-viz-qwen-gpqa}
\end{figure}

\begin{figure}[!ht]
\centering
\includegraphics[width=\textwidth]{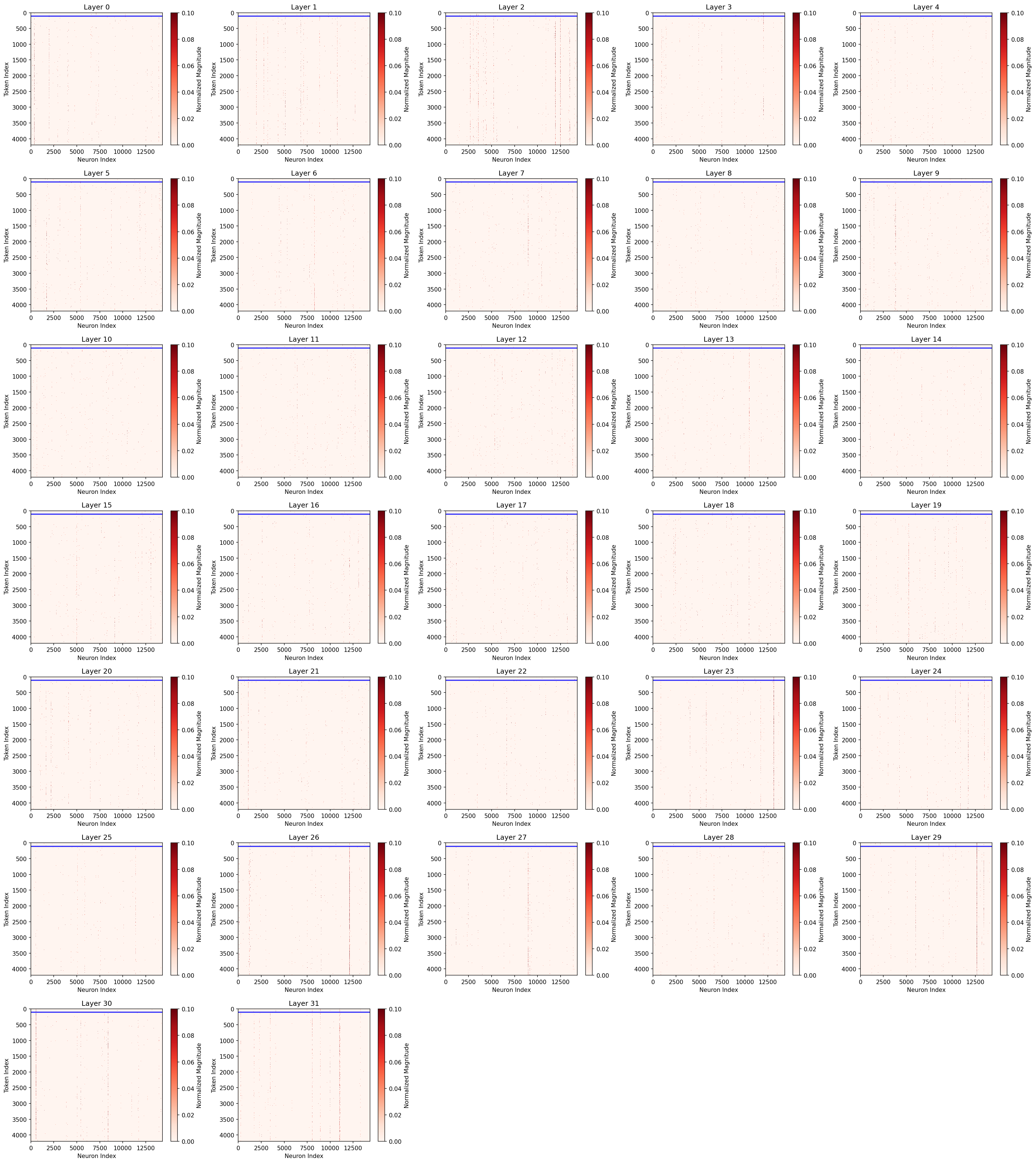}
\caption{Visualization of feedforward activations $\mathbf{Z}$ across all layers ($\bar{\mathbf{Z}} > 0.05$) of DeepSeek-R1-Distill-Llama-8B on MATH500. The x-axis represents neurons and the y-axis represents token indices from top to bottom. The blue horizontal line indicates the last position of the prompt.}
\label{fig:activation-viz-llama-math500}
\end{figure}

\begin{figure}[!ht]
\centering
\includegraphics[width=\textwidth]{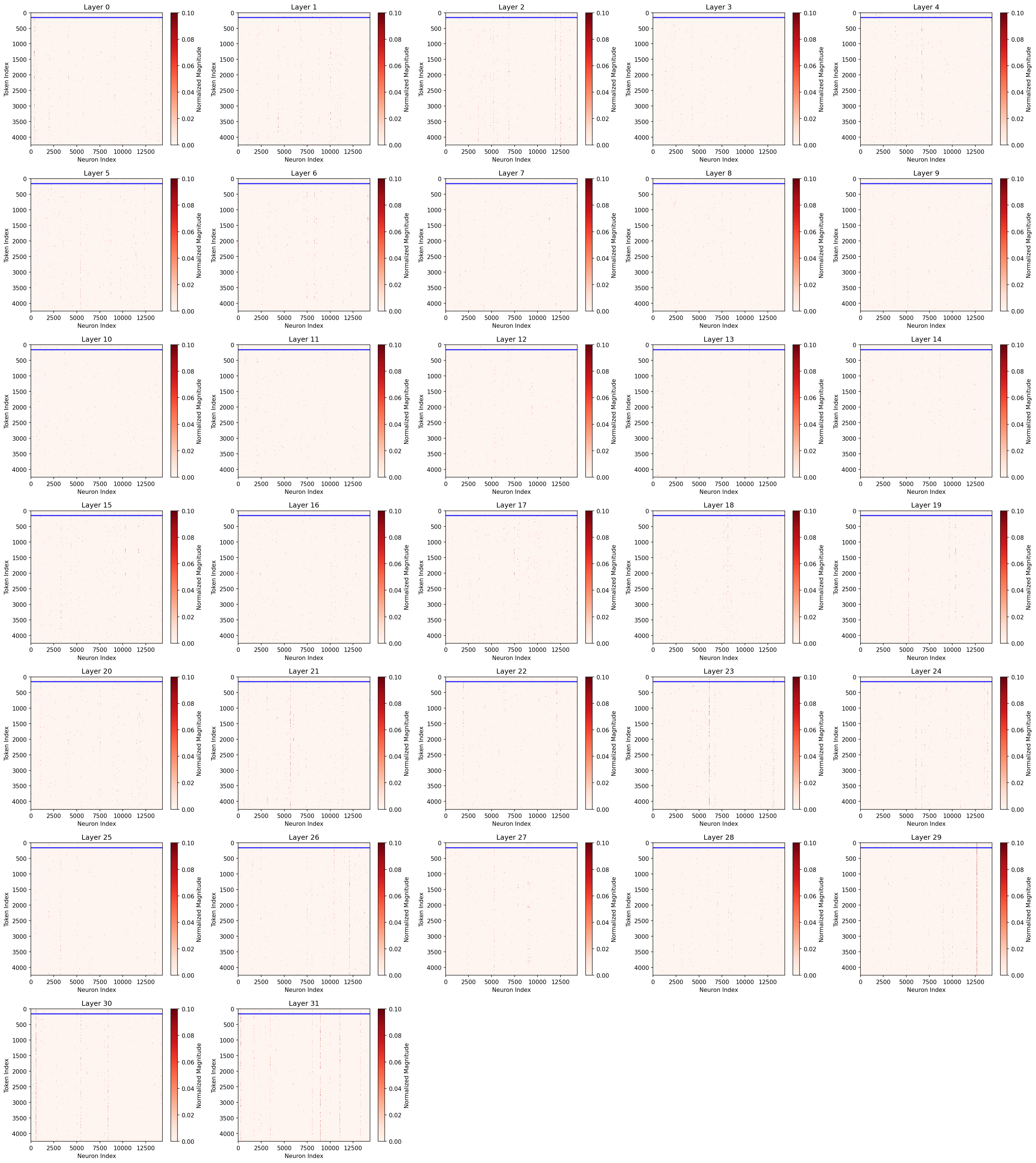}
\caption{Visualization of feedforward activations $\mathbf{Z}$ across all layers ($\bar{\mathbf{Z}} > 0.05$) of DeepSeek-R1-Distill-Llama-8B on GPQA-DIAMOND. The x-axis represents neurons and the y-axis represents token indices from top to bottom. The blue horizontal line indicates the last position of the prompt.}
\label{fig:activation-viz-llama-gpqa}
\end{figure}

Figures~\ref{fig:activation-viz-qwen-math500}--\ref{fig:activation-viz-llama-gpqa} visualize the feedforward activations $\mathbf{Z}$ across all layers for representative samples from MATH500 and GPQA-DIAMOND benchmarks (neurons with $\bar{\mathbf{Z}} > 0.05$ are shown). These visualizations further demonstrate the periodic re-firing behavior of important neurons during autoregressive generation, supporting the design of periodic mask updates in our method.

\FloatBarrier
\section{Use of Large Language Models}
\label{sec:llm-use}

We employed large language models (ChatGPT; ``GPT-5'' and Anthropic Claude) for English-language polishing, light copy-editing, and assisting with experimental code implementation. The models were not used to generate research ideas or experimental design. All technical content, claims, and experimental code were authored and verified by the authors to ensure correctness and reproducibility. No non-public data, confidential information, or personally identifiable information was provided to the models.

\end{document}